\documentclass[twoside,11pt]{article}

\usepackage[preprint]{jmlr2e_pja}

\usepackage{graphicx} 

\usepackage{xcolor}
\usepackage{multirow}
\usepackage{colortbl}

\usepackage{amsmath} 

\usepackage{makecell}

\newcommand{\sdash}{{\mbox{-}}}
\newcommand{\subtxt}[1]{{\mbox{\tiny #1}}}

\definecolor{issuePJA_color}{rgb}{1.0,0.0,0.0}

\definecolor{commentPJA_color}{rgb}{1.0,0.0,0.8}

\definecolor{rev_color}{rgb}{0.6,0.0,0.0}

\newcommand{\mb}[1]{\mathbf{#1}}
\newcommand{\bsy}[1]{\boldsymbol{#1}}

\definecolor{atz_table1}{rgb}{0.85, 0.85, 0.85}
\definecolor{atz_table2}{rgb}{0.8, 0.8, 0.8}
\definecolor{atz_table3}{rgb}{0.75, 0.75, 0.75}

\newcommand{\sminus}{\text{-}}
\newcommand{\splus}{\text{+}}

\jmlrheading{x}{xxxx}{xx-xx}{x/xx}{xx/xx}{xxxxxx}{Paul J. Atzberger}

\ShortHeadings{SDYN-GANs for Learning Stochastic Dynamics}{Atzberger}
\firstpageno{1}

\begin{document}

\title{SDYN-GANs: Adversarial Learning Methods for Multistep Generative Models for General Order Stochastic Dynamics}

\author{\name Panos Stinis \\
       {\addr Advanced Computing, Mathematics and Data Division\\
       Pacific Northwest National Laboratory (PNNL) \\
       Richland, WA} \\
       \\
       \name Constantinos Daskalakis \\
       {\addr Department of Electrical Engineering and Computer Science \\ 
       Massachusetts Institute of Technology (MIT) \\
       Cambridge, MA} \\
       \\
       \name Paul J.\ Atzberger {\email atzberg@gmail.com} \\
       {\addr Department of Mathematics\\
       Department of Mechanical Engineering \\
       University of California Santa Barbara (UCSB)\\
       Santa Barbara, CA 93106, USA} \\
       {\small \url{http://atzberger.org}}}

\maketitle

\begin{abstract}
We introduce adversarial learning methods for data-driven generative
modeling of the dynamics of $n^{th}$-order stochastic systems.  
Our approach builds on
Generative Adversarial Networks (GANs) with generative model classes based on
stable $m$-step stochastic numerical integrators.  
We introduce different
formulations and training methods for learning models of
stochastic dynamics based on observation of trajectory samples.  We develop 
approaches using discriminators based on Maximum Mean Discrepancy (MMD),
training protocols using conditional and marginal distributions, and methods
for learning dynamic responses over different time-scales.  We show how our approaches
can be used for modeling physical systems to learn force-laws, damping
coefficients, and noise-related parameters.  The adversarial learning 
approaches provide methods for obtaining stable generative models 
for dynamic tasks including long-time prediction and developing 
simulations for stochastic systems. 
\end{abstract}


\section{Introduction}

Learning and modeling tasks arising in statistical inference, machine learning,
engineering, and the sciences often involve inferring representations for
systems exhibiting non-linear stochastic dynamics.  This requires
estimating models that not only account for the observed behaviors but 
also allow for performing simulations or making predictions over longer time-scales.
Obtaining stable long-time simulations and predictions requires data-driven 
approaches capable of learning structured dynamical models from observations.

Estimating parameters and representations for stochastic dynamics has a
long history in the statistics and engineering
community~\citep{Peeters2001,Hurn2007,Jensen2002,Kalman1960,Nielsen2000}.
This includes methods for inferring models and
parameters in robotics~\citep{Olsen2002}, orbital
mechanics~\citep{Muinonen1993}, geology~\citep{PardoIguzquiza1998}, political
science~\citep{King1998}, finance~\citep{Kim1998}, and
economics~\citep{Cramer1989}. 
One of the most prominent strategies for developing estimators is to
use Maximum Likelihood Estimation (MLE)~\citep{Fisher1922,Hald1999} or Bayesian
Inference (BI)~\citep{Gelman1995}.  For systems where likelihoods can be 
specified readily and computed tractably, estimators can be developed 
with asymptotically near optimal data 
efficiency~\citep{Book_All_of_Statistics_Wasserman_2010}.
However, in general, specifying and computing likelihoods for MLE, especially
for stochastic processes, can pose significant challenges in 
practice~\citep{Hurn2007,Sorensen2004}.  For Bayesian methods, 
this is further compounded by well-known challenges in computing
and analyzing posterior distributions~\citep{Gelman1995,Smith1993}.  
For stochastic systems, this is also compounded by the high dimensionality 
of the ensemble of observations and data consisting of 
trajectories of the system.

As an alternative to MLE methods, we develop adversarial learning methods using
discriminators/critics based on a generalization of the method-of-moments
building on~\citep{Hansen1982,Gretton2012,Fortet1953}.  We develop methods for
learning models and estimating parameters using a generative modeling approach
which compares samples of candidate models with the observation data.  A
notable feature of these sample-based adversarial methods is they is no longer
require specification of likelihood functions.  This allows for learning over
more general classes of generative models.  The key requirement now becomes
identifying a collection of features of the samples 
that are sufficiently rich for the discriminators/critics
to make comparisons with the observation data for model selection.
We develop methods based on characteristic 
kernels and generalized moments for distinguishing general classes of 
distributions~\citep{Gretton2012,Fortet1953}.

We introduce adversarial learning approaches and training methods 
for obtaining representations of
$n^{th}$-order stochastic processes using a class of generative models 
based on $m$-step stochastic numerical methods.  
For gradient-based learning with loss functions based on 
second-order statistics, and the high dimensional samples
arising from trajectory observations, we develop specialized 
adjoint methods and custom backpropagation methods.  Our 
approaches provide Generative Adversarial Networks (GANs) for
learning representations of the non-linear dynamics of
stochastic systems.  We refer to the methods 
as Stochastic-Dynamic Generative Adversarial Neural 
Networks (SDYN-GANs).

\section{Adversarial Learning}

\begin{figure}[ht]
\centerline{\includegraphics[width=0.65\columnwidth]{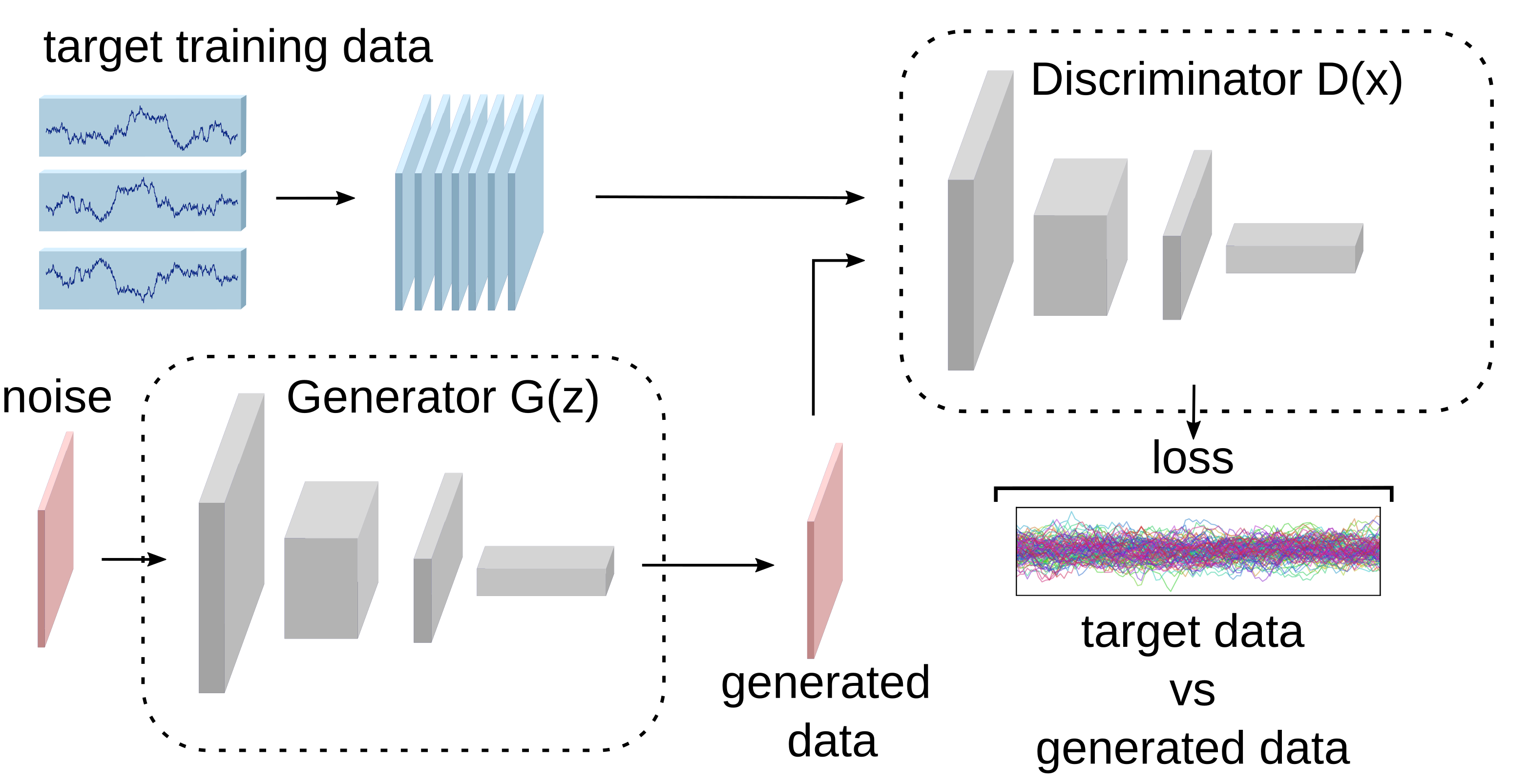}}
\caption{\textbf{Adversarial Learning.} A generative model $G(\cdot;\theta_G)$ and 
discriminator/critic $D(\cdot;\theta_D)$ are used to learn models from 
samples of the trajectory data.  The 
$D$ is used in computing a loss comparing 
samples $\{\mb{\tilde{x}}_i\}_{i=1}^{\tilde{N}}$ generated by $G$ with the samples 
$\{\mb{x}_i\}_{i=1}^N$ of the training data. \label{fig_adv_learning} 
\label{fig_adv_learning_schematic}}
\end{figure}

We develop adversarial learning approaches for the purpose of quantifying the
differences between a collection of samples from candidate generative models
from those of the training data.  In adversarial learning a generative model
$G(\cdot;\theta_G)$ is learned in conjunction with a discriminative model
$D(\cdot;\theta_D)$.  This can be viewed as a two-player game.  The aim of the
discriminator $D$ is to minimize errors in distinguishing a collection of samples
of the generator from those of the training data.  The aim of the generator is to
produce samples of such good quality that the discriminative model is confused
and would commit errors in being able to distinguish them from the training data,
see Figure~\ref{fig_adv_learning_schematic}.
Depending on the choices made for the loss functions, and the model classes of
the discriminative and generative models, different types of divergences and
metrics can be obtained.  These choices impact how comparisons are made between
the distribution of the generative model
with the distribution of the training data.  In practice, learning methods are
sought that perform well using only samples from these distributions. 

\subsection{Generative Adversarial Networks (GANs)}
\label{sec_GANs}

In the adversarial training of GANs, the first model $G$ is a neural network
representing an implicit generative model denoted by $G(\mb{Z};\theta_G)$ that
generates samples from a random source $\mb{Z}$.  The $\theta_G$ are the
parameters of the model and $\mb{Z}$ denotes a random variable mapped under $G$
to generate samples $\tilde{\mb{x}}^{[i]} = G(\mb{Z}^{[i]};\theta_G)$.  The
second model $D$ serves to distinguish how well a set of samples
$\{\tilde{\mb{x}}^{[i]}\}_{i=1}^{\tilde{N}}$ generated by $G$ matches samples
$\{\mb{x}^{[i]}\}_{i=1}^{N}$ with those in the training data set. Let the empirical
distributions be denoted for the generator by $\tilde{\mu}_G$ and for the
training data by $\tilde{\mu}_\mathcal{D}$.  The discriminator/critic model is
denoted by $D(\cdot;\theta_D)$.  The choices for the form of $D$ and for the
loss functions $\ell$ determine how the comparisons are made between the
empirical probability distributions of
the two sets of samples, $\tilde{\mu}_G,\tilde{\mu}_\mathcal{D}$. 
There are many ways to formulate GANs and related
statistical estimators~\citep{Sriperumbudur2009a,Goodfellow2016}. 

In the original formulation of GANs, the objective can be expressed 
as \\
$\theta_D^* = \arg\min_{\theta_D} J_D(\theta_G,\theta_D)$
with  
$J_D = \mathbb{E}_{(x,y)\sim p_{\subtxt{synth}}}
\left[
-\log p_\subtxt{D}(y|x;\theta_D)
\right]$,~\citep{Goodfellow2014,Goodfellow2016}.  
The $p_{\subtxt{synth}}(x,y) = \frac{1}{2} p_\subtxt{data}(x,y) 
+ \frac{1}{2}p_\subtxt{model}(x,y;\theta_G)$ 
is the probability
of a synthetic data distribution obtained by mixing the samples of the
generator $\tilde{x} \sim p_\subtxt{model} \sim \tilde{\mu}_G$ with the
training data $x \sim p_\subtxt{data} \sim
\tilde{\mu}_{\mathcal{D}}$,~\citep{Goodfellow2016}.  The labels are assigned to
the samples with $y=-1$ for the generated data
samples and $y=+1$ for training data.  For a given $\theta_D$, we let $D(x) =
p_\subtxt{D}(y=+1|x;\theta^D)$ for the probability that the discriminator
classifier assigns the sample $x$ to have the label $y=+1$.  We then have $1 - D(x) =
p_\subtxt{D}(y=-1|x;\theta_D)$.  
With these choices and notation, the objective now can be 
expressed as $J_D = -\frac{1}{2}
\mathbb{E}_{x\sim p_{\subtxt{data}}}
\left[
\log(D(x))
\right]
-\frac{1}{2}
\mathbb{E}_{x\sim p_{\subtxt{model}}(;\theta_G)}
\left[
\log(1 - D(x))
\right].
$
The aim is
for the generator to produce high quality samples so that even
the optimal discriminator $\theta_D^*$ is  
confused and gives the value $D(x) = p_\subtxt{D}(\cdot;\theta_D^*) = 1/2$. 

The optimal generative model in this case has objective
$\theta_G^* = \arg \min_{\theta_G} J_G(\theta_G,\theta_D^*(\theta_G))$.
When $J_G = -J_D$, this can be viewed as a two-player zero-sum game with cost function $\ell = J_D$ 
with $\theta_D^*$ and $\theta_G^*$ corresponding to the optimal strategies. 
If the objectives are fully optimized to the equilibrium, this would give 
the Jensen-Shannon (JS) Divergence
$JS(\tilde{\mu}_{G(;\theta_G)},\tilde{\mu}_\mathcal{D})$~\citep{Lin1991}
for comparing the empirical 
distributions of the generator $\tilde{\mu}_G$ and training 
data $\tilde{\mu}_\mathcal{D}$.  The optimal generator would be 
$\theta_G^* = \arg\min_{\theta_G}
JS(\tilde{\mu}_{G(;\theta_G)},\tilde{\mu}_\mathcal{D})$, where
$JS(\tilde{\mu}_G,\tilde{\mu}_\mathcal{D}) = \frac{1}{2}D_{KL}\left(\tilde{\mu}_\mathcal{D}
\| \frac{1}{2}\left(\tilde{\mu}_G + \tilde{\mu}_\mathcal{D}  \right)  \right) +
\frac{1}{2}D_{KL}\left(\tilde{\mu}_G  \| \frac{1}{2}\left(\tilde{\mu}_G
+ \tilde{\mu}_\mathcal{D}  \right)  \right) $, with $D_{KL}$ the Kullback-Leibler 
Divergence~\citep{Book_Cover_Thomas_2006,Kullback1951}.
In practice, to mitigate issues 
with gradient-learning, the zero-sum condition can 
be relaxed by using different loss functions in the objectives
with $J_G \neq -J_D$.  For example, $J_G = 
-\frac{1}{2}
\mathbb{E}_{x\sim p_{\subtxt{model}}(;\theta_G)}
\left[
\log(D(x))
\right]$
is used sometimes for the generator $G$ to improve the
gradient~\citep{Goodfellow2016}.  Such formulations 
can result in other types of games with
more general solutions characterized as Nash Equilibria or other stationary
states~\citep{Nisan2007,Owen2013,Nash1951,Roughgarden2010}.

GANs and other statistical estimators 
also can be formulated using discriminators/critics based on 
$J_D(\theta_G,\theta_D) =  
\mathbb{E}_{\tilde{\mu}_{G(;\theta_G)}}\left[F(X';\theta_D)
\right]
-\mathbb{E}_{\tilde{\mu}_\mathcal{D}}\left[F(X;\theta_D)
\right]
$ and $J_G = -J_D$ with \\ $\theta_G^* = \arg\min_{\theta_G} J_G(\theta_G,\theta_D^*(\theta_G))$
and $\theta_D^* = \arg\min_{\theta_D} J_D(\theta_G,\theta_D)$.
This objective can be viewed from a statistical perspective
as yielding an Integral Probability Metric
(IPM)~\citep{Sriperumbudur2009a,Muller1997}.  The IPM can be expressed as \\ 
$ 
\theta_G^* = \arg\min_{\theta_G}\max_{f \in \mathcal{F}}
\left(
\mathbb{E}_{\tilde{\mu}_\mathcal{D}}\left[f(X)
\right]
-\mathbb{E}_{\tilde{\mu}_{G(;\theta_G)}}\left[f(X')
\right]
\right)
= \arg\min_{\theta_G}\max_{f \in\mathcal{F}} \ell(\theta_G;f) = \arg\min_{\theta_G} \ell^*(\theta_G)
$, where $\mathcal{F} = \{f \;|\; \exists \theta_D \mbox{ s.t. } f = F(\cdot;\theta_D)\}$
and $\ell^*(\theta_G) = \max_{f \in \mathcal{F}} \ell(\theta_G;f)$.
In the case of $\mathcal{F} = \mbox{Lip-}1$, consisting of Lipschitz functions
with $L \leq 1$, this results in $\ell^*(\theta_G) = \max_{f \in {\small
\mbox{Lip-}1}} \ell(\theta_G;f)=
\mathcal{W}^1\left(\tilde{\mu}_G,\tilde{\mu}_{\mathcal{D}} \right)$. This
follows by Kantorovich-Rubinstein Duality, where $\mathcal{W}^1$ is the
Wasserstein-1 distance for the empirical measures $\tilde{\mu}_G$ and
$\tilde{\mu}_\mathcal{D}$~\citep{Villani2008}. While Wasserstein distance provides a
widely used norm for analyzing measures, in practice for training, it can be computationally
expensive to approximate and can result in challenging gradients for
learning~\citep{Mallasto2019}. 

As an alternative to these approaches, we use for IPMs a smoother class of discriminators, 
with
$\mathcal{F} \subset \mathcal{H}$ with $\mathcal{H}$ a Reproducing Kernel
Hilbert Space (RKHS)~\citep{Aronszajn1950,Berlinet2011}. By choice of the
kernel $k(\cdot,\cdot)$ we also can influence the features of the distributions
that are emphasized during comparisons. We take $\mathcal{F} = \{f \in
\mathcal{H} \; | \; \|f\|_\mathcal{H} \leq 1 \}$ for the objective,
$\theta_G^* = \arg\min_{\theta_G}\max_{f \in \mathcal{F}}
\ell(\theta_G;f)$, where 
$\ell(\theta_G;f) = \mathbb{E}_{\tilde{\mu}_\mathcal{D}}\left[f(X)
\right]
-\mathbb{E}_{\tilde{\mu}_{G(;\theta_G)}}\left[f(X')
\right]
$.

The objective can be partially solved analytically to yield
$\ell^*(\theta_G) =
\max_{f \in \mathcal{F}} \ell(\theta_G;f) = 
\|\eta_{\mathbb{P}} - \eta_{\mathbb{Q}} \|_{\mathcal{H}}$, where
$\eta_{\mathbb{P}} = E_{X \sim \mathbb{P}}\left[k(\cdot,X)\right]$,
$\eta_{\mathbb{Q}} = E_{X' \sim \mathbb{Q}}\left[k(\cdot,X')\right]$ and $\|\cdot
\|_{\mathcal{H}}$ is the norm of the RKHS~\citep{Sriperumbudur2010}. 
Here, we take 
$\mathbb{P} = \tilde{\mu}_\mathcal{D}$ to be the empirical distribution of the 
training data and $\mathbb{Q} = \tilde{\mu}_G$ the empirical distribution of the
generated samples. Characterizing the
differences between the probability distributions in this way corresponds to
the statistical approaches of Generalized Method of Moments~\citep{Hansen1982}
and Maximum Mean Discrepancy
(MMD)~\citep{Fortet1953,Smola2006}. 

\subsection{SDYN-GANs Adversarial Learning Framework for Stochastic Dynamics}
\label{sec_sdyn_gans}
\label{sec_training_protocols}

We discuss our overall adversarial training approach for learning generative
models for stochastic dynamics, and give more technical details in the later
sections. Learning representations of stochastic dynamics capable of making
long-time predictions presents challenges for direct application of traditional
estimation methods not taking temporal aspects into account.  
This arises from the high dimensionality of trajectory
data samples and in the need for stable models that can predict beyond the time
duration of the observed dynamics.  To obtain reliable learning methods
requires making effective use of the special structures of the 
stochastic system and trajectory data.  This includes
Markovian conditions, probabilistic dependencies, and 
other properties. 

While our approaches can be used more broadly, we focus here on
learning generative models related to vector-valued Ito Stochastic
Processes~\citep{Oksendal2000}.  This is motivated by systems with
unknown features which can be modeled by solutions of
Stochastic Differential Equations (SDEs)~\citep{Oksendal2000,Gardiner1985}. 
From a
probabilistic graphical modeling
perspective~\citep{Wainwright2008,Bishop_Book_ML_2006,Sucar2015,Murphy2012}, 
the high dimensional
probability distributions associated with observations of trajectories can 
be factored.  This allows for dependencies to be used to obtain more 
local models coupling the components over time greatly reducing the 
dimensionality of the problem.  For
obtaining generative models $G$ that are useful for prediction and long-time
simulations, we develop generator classes based on stable $m$-step stochastic
numerical integrators with learnable components~\citep{Platen1992}. 
As we discuss in later sections, 
we motivate and investigate the performance of our methods by considering problems
for physical systems and mechanics in $\mathbb{R}^3$.  In this case, we can further refine
the class of numerical discretizations to incorporate physical principles,
such as approximate time-reversibility for the conservative contributions to
the dynamics, fluctuation-dissipation balance, and other 
properties~\citep{Verlet1967,Farago_Langevin_Num_Method_2013,
Jasuja2022,AtzbergerTabak2015,TraskAtzberger_GMLS_Nets_AAAI_2020,Lopez2022}.   

\begin{figure}[ht]
\centerline{\includegraphics[width=0.89\columnwidth]{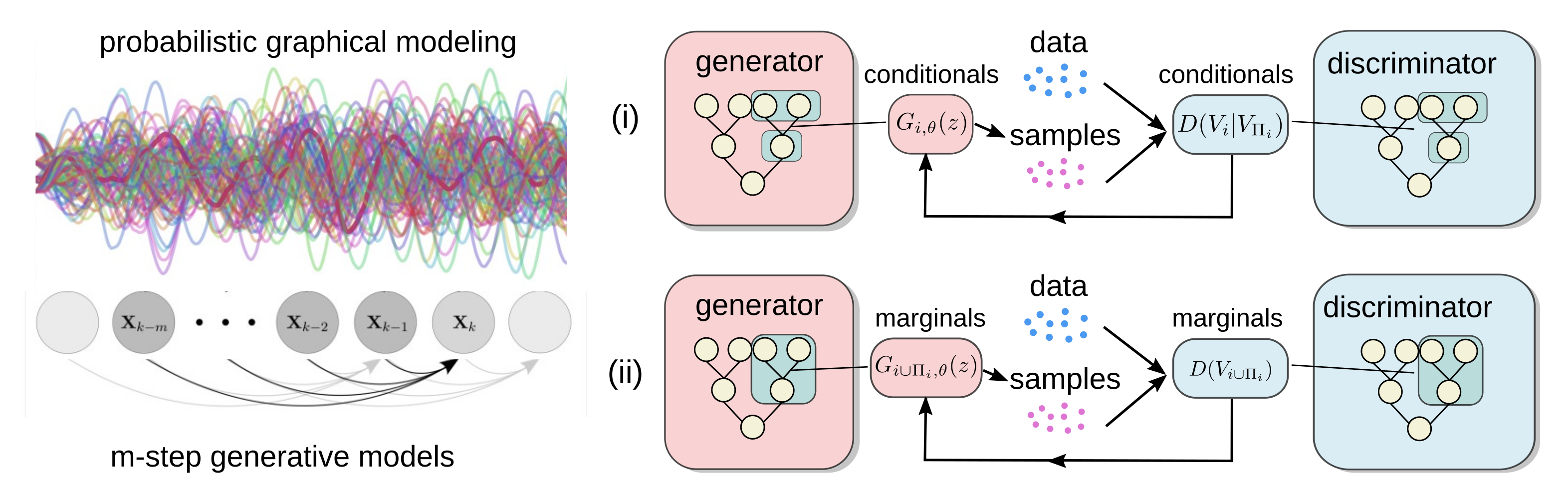}}
\caption{\textbf{Decompositions for Training.} 
We utilize for the trajectories the probabilistic dependence between the configurations $\mb{X}(t)$ 
to help facilitate learning models from the time series data \textit{(left)}.  We develop different training protocols using (i) conditional statistics 
and (ii) marginal statistics.  We also develop training methods for different  
trajectory durations $[0,\mathcal{T}]$ and sub-sampling 
over time-scales $\tau$ \textit{(right)}. }
\label{fig_graphical_model}
\end{figure}

We develop Stochastic Dynamic Generative Adversarial Networks (SDYN-GANs), 
for learning generative models $G$  
for sampling discretized trajectories, 
$\mb{X}^{[i]} = G(\mb{Z}^{[i]},\theta_G)$.  The $\mb{Z}^{[i]}$ 
denotes the sample of the noise source mapped under $G$ to 
generate the sample $\mb{X}^{[i]}$.
The data samples 
can be expressed as 
$\mb{X}^{[i]} = \left(\mb{X}_1^{[i]},\ldots,\mb{X}_N^{[i]} \right)$
where $\mb{X}_j^{[i]}=\mb{X}^{[i]}(t_j)$ are the temporal components.  We learn 
generators $G$ having the form of $m$-step stochastic numerical integration methods 
$\mb{X}_j = \bsy{\Psi}_{j-1}(\mb{X}_{j-1},\ldots,\mb{X}_{j-m},\bsy{\omega}_{j-1};\mb{p})$
with initial conditions $\mb{X}_0 = \mb{x}_0(\mb{p}),\ldots,\mb{X}_{m-1} = \mb{x}_{m-1}(\mb{p})$.
The trainable term
$\bsy{\Psi}_{j-1}$ approximates the evolution of the stochastic dynamics from
time-step $t_{j-1}$ to $t_j$ using information at times
$t_{j-1},\ldots,t_{j-m}$. The numerical updates
are randomized with samples $\bsy{\omega}_{j-1}$ and have learnable parameters 
$\mb{p}$.  
In terms of $G(\mb{Z};\theta_G)$, we will consider here primarily the case 
with $\mb{Z} = \{\bsy{\omega}_j\}_{j}$ and 
$\theta_G = \mb{p}$.  In the general case, 
$\mb{Z}$ can also include additional 
sources of noise and $\theta_G$ additional parameters.
The use of $m$-step stochastic methods for our generative modeling 
allows for leveraging probabilistic dependencies to factor high dimensional
trajectory data.  Our $m$-step methods can be viewed as a variant of structural
equations for  probabilistic graphical
modeling~\citep{Bishop_Book_ML_2006,Murphy2012,Wainwright2008,Daskalakis_Subadditive_Graphs_2021},
see Figure~\ref{fig_graphical_model}.

We use adversarial learning to train the generative model $G$ by using discriminators/critics $D$ 
to distinguish the samples generated by $G$ from those of the training data. 
We obtain generators by optimizing the loss
$\theta_G^* = \arg \min_{\theta_G} \ell^*(\theta_G)$, with $\ell^*(\theta_G) = \ell(\theta_G,
\theta_D^*(\theta_G))$.
This can be formulated many different ways corresponding to different
divergences and metrics of the empirical distributions as we discussed in
Section \ref{sec_GANs}.  We use losses $\hat{\ell}^*$ corresponding to Integral
Probability Metrics (IPMs) of the form
$\hat{\ell}^*(\theta_G) = \hat{\ell}^*(\mathbb{P},\mathbb{Q}) 
= \max_{f \in \mathcal{F}}\left( 
\mathbb{E}_\mathbb{P}\left[f(\mb{X})\right] - 
\mathbb{E}_\mathbb{Q}\left[f(\mb{X'})\right]\right)$,
where $\mathbb{P}$, $\mathbb{Q}$ denote the 
measures being compared.  We take 
$\mathbb{P} = \tilde{\mu}_\mathcal{D},\mathbb{Q} = \tilde{\mu}_G$
and use 
$\mathcal{F} = \{f \in \mathcal{H} \;|\; \|f\|_{\mathcal{H}} \leq 1\}$,
where $\mathcal{H}$ is a Reproducing Kernel Hilbert Space (RKHS) with kernel 
$k(\cdot,\cdot)$. 
With these choices, we can express the GANs and resulting IPM loss as
$\hat{\ell}^*(\theta_G) = \|\eta_{\mathbb{Q}} - \eta_{\mathbb{P}}
\|_{\mathcal{H}}$, where $\eta_{\mathbb{P}} =
\mathbb{E}_{\mathbb{P}}\left[k(\cdot,\mb{X})\right]$ and $\eta_{\mathbb{Q}} =
\mathbb{E}_{\mathbb{Q}}\left[k(\cdot,\mb{X})\right]$.  This can be viewed as
embedding each of the measures as functions in the RKHS
$\mathcal{H}$~\citep{Saitoh2016,Sriperumbudur2010}.  The characterization of
their differences can be expressed as the $\mathcal{H}$-norm of their embedded
representations.  Since $k(\cdot,\mb{X})$ gives a collection of features, which
under expectation generate generalized moments, this approach corresponds to 
Generalized Moments Methods (GMMs)~\citep{Hansen1982} and Maximum Mean
Discrepancy (MMD)~\citep{Gretton2012,Fortet1953,Smola2006}.  

In practice, to obtain more amenable losses for optimization and for computing gradients, we use a 
reformulation which
squares the objective to obtain $\ell^*(\theta_G) = \left(\hat{\ell}^*(\theta_G)\right)^2 =
\|\eta_{\mathbb{P}} - \eta_{\mathbb{Q}} \|_{\mathcal{H}}^2 =
\mathbb{E}_{\mb{X}\sim\mathbb{P}}\left[k(\mb{X},\mb{X}) \right] -2
\mathbb{E}_{\mb{X}\sim\mathbb{P},\mb{X'}\sim\mathbb{Q}}\left[k(\mb{X},\mb{X'}) \right] +
\mathbb{E}_{\mb{X'}\sim\mathbb{Q}}\left[k(\mb{X'},\mb{X'}) \right]$.  This is related to
energy statistics~\citep{Szekely2013a} and the sample-based estimator
developed in~\citep{Gretton2012}.
To find approximate minimizers for training $G$, we use Stochastic
Gradient Descent (SGD) methods 
$\theta_G^{n+1} = \theta_G^{n} - \eta_n \hat{q}$,
with learning rate
$\eta_n$ and gradient estimators 
$\hat{q} \approx
\nabla_{\theta_G} \ell^*(\theta_G^n)$ based on mini-batches
~\citep{RobbinsMonro1951,Bottou1991,Bishop_Book_ML_2006}.  We 
use variants incorporating momentum, such as ADAM and 
others~\citep{Kingma2014,Tieleman2012,Lydia2019}.  To compute
$\hat{q}$, we use the unbiased estimator of~\citep{Gretton2012} as part 
of the gradient approximation.  We estimate gradients for
the empirical loss using 
$
\hat{q} = \nabla_{\theta_G} \tilde{\ell}^*(\theta_G)
$
where
\begin{eqnarray}
\label{equ_tilde_ell_star}
\tilde{\ell}^*(\theta_G) =  \hspace{0.91\columnwidth} &&\\ 
\nonumber
\frac{1}{N(N-1)} \sum_{i\neq j}^{N} 
k\left(\mb{X}^{[i]},\mb{X}^{[j]}\right)
- \frac{2}{N M} \sum_{i=1}^{N} \sum_{j=1}^{M} 
k\left(\mb{X}^{[i]},\mb{Y}^{[j]}\right) 
+ 
\frac{1}{M(M-1)}\sum_{i\neq j}^{M} 
k\left(\mb{Y}^{[i]},\mb{Y}^{[j]}\right).
\end{eqnarray}
The samples of mini-batches have distributions
$\mb{X}^{[i]} \sim \tilde{\mu}_G$ 
and $\mb{Y}^{[i]} \sim \tilde{\mu}_\mathcal{D}$.

The trajectory samples $\mb{X}^{[i]}=G(\mb{Z}^{[i]};\theta_G)$ depend on
$\theta_G$, which further requires methods for computing the contributions of
the gradients $\nabla_{\theta_G} \mb{X}^{[i]}$. This can pose
challenges since these gradients involve the $m-$step stochastic integrators when $\mb{p}
\subset \theta_G$.  This can become particularly expensive if using direct
backpropagation based on book-keeping for forward evaluations of the stochastic integrators
and the second-order statistics of $\tilde{\ell}^*$.  This would result in
prohibitively large computational call graphs.  This arises from the high
dimension of the trajectories $\mb{X}$ and the multiple dependent steps
during generation.  This is further compounded by the mini-batch sampling
$\{\mb{X}^{[i]}\}_{i=1}^{n_b}$ and second-order statistics.  As we discuss in
the later Sections~\ref{sec_m_step_gradients} and~\ref{sec_backprop_loss}, we
can address these issues by developing adjoint methods and custom
backpropagation approaches specialized to utilize the structure of the loss
functions.  This allows us to compute efficiently the contributions of
$\nabla_{\theta_G} \mb{X}^{[i]}$ to $\nabla_{\theta_G} \tilde{\ell}^*$ to
obtain the gradient estimates needed for training. 

We also introduce and investigate for SDYN-GANs training protocols utilizing
the trajectory data in a few different ways to compare the underlying
probability distributions, see Figure~\ref{fig_graphical_model}.  This includes
comparisons using (i) distributions over the entire trajectory, (ii) marginal
distributions over trajectory fragments, and (iii) conditional distributions
over trajectory fragments. We develop the following training approaches for
$n^{th}$-order stochastic systems
\begin{itemize}
 \item[] \textbf{\textit{Full-Trajectory Method (full-traj)}}: Samples full trajectory 
  $\mb{R} = (\mb{X}(t_0),\mb{X}(t_0 + \delta{t}),
  \ldots,\mb{X}(t_0 + (m-1)\delta{t}),
  \mb{X}(t_1 ),
  \mb{X}(t_1 + \tau),\ldots,\mb{X}(t_1 + (N_\subtxt{$\mathcal{T}$}-1)\tau))$
  corresponding to distribution $p_\theta(\mb{R})$ and trajectory with $N_\subtxt{$\mathcal{T}$}$ steps. 
 \item[]\textbf{\textit{Marginals Method (marginals)}}: Samples marginals over 
 trajectory fragments $\mb{R}_k$ of the form
 $\mb{R}_k =
  (\mb{X}(t_k),\mb{X}(t_k + \delta{t}),
  \ldots,\mb{X}(t_k + (m-1)\delta{t}),
  \mb{X}(s_k),\mb{X}(s_k + \tau),\ldots,\mb{X}(s_k + (\ell-1)\tau))$, $k \in [1,N_m]$,
with marginal distributions
  $\{p_{\theta,k}(\mb{R}_k)\}_{k=1}^{N_m}$.
 \item[]\textbf{\textit{Conditional Method (conditionals)}}: Samples 
 conditionals $\mb{R}_k | \mb{Q}_k$ over 
  trajectory fragments $\mb{R}_k$ conditioned on 
  fragments $\mb{Q}_k$ of the form
  $\mb{R}_k
  = (\mb{X}(s_k),\mb{X}(s_k + \tau),\ldots,\mb{X}(s_k + (\ell-1)\tau))$ and 
  $\mb{Q}_k 
  = (\mb{X}(t_k),
  \mb{X}(t_k + \delta{t}),
  \ldots,\mb{X}(t_k + (m-1)\delta{t})$, $k \in [1,N_m]$, with conditional distributions
  $\{p_{\theta,k}(\mb{R}_k|\mb{Q}_k)\}_{k=1}^{N_m}$.
\end{itemize}

In each of these approaches the $t_k$ is the time at which the current state of
the $n^{th}$-order dynamical system is represented in the training data. The
$\delta{t}$ corresponds to the time-scale for sampling the initial dynamical state
information. The $s_k$ give the part of the trajectory on which we seek to
predict the dynamics. The $\tau$ is the time-scale on which the evolving
dynamics are sampled for training and prediction. In
practice, we typically will have $n=m$, however, even when $n=1$ our 
formulation allows for $m > 1$ for use of stochastic multistep 
numerical methods, such as Adams-Bashforth~\citep{Platen1992,Buckwar2006,Ewald2005}.

In practice, typical choices are to have $\delta{t}=\Delta{t}$, and
$\tau=a\Delta{t}$ for $a \in \mathbb{N} $ to obtain aligned temporal sampling
between the stochastic trajectories of the generative models and the
observation data.  We will take as our default choice $t_k = k\Delta{t}$ and
$s_k = t_k + \tau$.  In principle, these also can be chosen more generally
provided a method for interpolation is given for comparing the temporal samples
between the training data and the generative models.  We discuss additional
technical details in the sections below.  We investigate SDYN-GANs using these
different training protocols (i)--(iii) for learning physical model parameters
and force-laws in Section~\ref{sec_results}.  

\section{Related Work}

There have been previous methods introduced for training GANs to model
deterministic and stochastic dynamics, including
Neural-ODEs/SDEs~\citep{Duvenaud_Neuro_ODEs_2018,Kidger2021} and motivated by
problems in physics~\citep{Darve_GANs_2022,AvilaBelbutePeres2018,Stinis2019,
Karniadakis_Yang_Phys_Informed_SDEs_2020,YangDaskalakisKarniadakis2022}.
Recent GANs methods are closely related to prior adjoint and system
identification methods developed in the engineering
communities~\citep{Lions1971,Giles2000,Ljung1987,Cea1986} and finance
communities~\citep{Kaebe2009,Giles2006}.  In these GANs methods, Maximum
Likelihood Estimation (MLE) is replaced with either the initial GANs
approximation to Jensen-Shannon (JS) metric by using neural network classifier
discriminators~\citep{Goodfellow2014} or by approximations to the Wasserstein
metric~\citep{WGANs_Orig_Bottou_2017} based on neural networks with Gradient
Penalties (WGAN-GP)~\citep{WGAN_GP_A_Courville_2017} or with Wasserstein
metrics for marginal slices~\citep{Deshpande2019}.  For time series analysis,
probabilistic graphical models also have been combined with GANs to enhance
training and characterized by establishing subadditivity bounds
in~\citep{Daskalakis_Subadditive_Graphs_2021,Daskalakis_Subadditive_2020}.

The use of general neural network discriminators are known to be challenging to
train for GANs.  This manifests as oscillations during training with well-known
issues with convergence and stability~\citep{Kodali2017,ThomasPinetz2018}.
While Neural ODEs/SDEs have been introduced for gradient
training~\citep{Duvenaud_Neuro_ODEs_2018,Tzen2019}, they make approximations in
the gradient calculations which can result in inconsistencies with the chosen
discretizations representing the stochastic dynamics.  These methods also treat
dynamics primarily of first-order systems.  

In our methods, we
develop a set of adjoint methods for computing gradients for general
$n^{th}$-order dynamics where it is particularly important for stability and
consistency during training to explicitly take into account the choice of the
$m$-step discretization methods used.  In our methods, this ensures our
training gradients align with our specified loss functions for our structured
$m$-step generative model classes.  While our adversarial learning framework
can be used generally, we mitigate oscillations and other issues in
training by focusing on using $D$ where the discriminators are 
analytically tractable.  Our $D$ are based on a collection of 
feature maps which can be viewed as employing kernel methods 
to encode and embed the probability distributions 
into a  Reproducing Kernel Hilbert Space
(RKHS)~\citep{Gretton2012,Sriperumbudur2010,Saitoh2016}.  The comparisons
between the samples of the candidate models and the observation data correspond to a
set of losses related to Generalized Moments Methods (GMMs)~\citep{Hansen1982}
and Maximum Mean Discrepancy (MMD)~\citep{Gretton2012,Fortet1953}.

Our methods with this choice for losses is related to GANs-MMD approaches that
previously were applied to problems in image
generation~\citep{Li2015,Gretton2018,Ghahramani_MMD_2015}.  Our work addresses
generative models for $n^{th}$-order stochastic dynamics posing new 
challenges arising from the temporal aspects, high dimensional trajectory samples, 
and higher-order dynamics.  Related work
using MMD to learn SDE models was developed in~\citep{Scholkopf_SDEs_2019}. In
this work, the focus was on scalar-valued first-order SDEs and learning the
drift term by using a splitting of the dynamics and using Gaussian Process
Regression (GPR)~\citep{GPR_Williams_Rasmussen_1996}.  The GPR was used as a
smoother to develop estimators for the drift contributions. For their constant
covariance term, they learned the proportionality constant by treating it as part
of the hyperparameters for the prior distribution for the GPRs optimized by
MLE~\citep{Scholkopf_SDEs_2019}.  

In our work, we develop methods for treating more general vector-valued
$n^{th}$-order stochastic dynamics and learning simultaneously the
contributions of both the drift and diffusion.  Motivated by problems in
microscale mechanics, we show how our methods can be used to learn
representations of vector-valued second-order inertial stochastic dynamics.  We
also introduce structured classes of implicit generative models by building on
$m$-step stochastic numerical methods.  

For training models, we
develop adjoint methods to obtain gradients consistent with 
the specific choices of the $m$-step stochastic
discretizations employed.  We also develop custom backpropagation
procedures for taking into account the second-order statistics of the losses.
Given the intrinsic high dimensionality of trajectory data, we also develop
training methods for reducing the dimension by sub-sampling trajectories over
time-scales $\tau$. 
Our learning methods allow for adjusting the time-scales
overwhich training is responsive to the observation data to facilitate methods
for temporal filtering and coarsening.  We also develop different training
protocols with statistical criteria based on the conditional distributions or
marginal distributions of the stochastic dynamics.  In summary, our methods
provide flexible sample-based approaches for training over general classes of
generative models to learn structured representations of the $n^{th}$-order 
dynamics of stochastic systems. 

\section{Learning Generative Models with SDYN-GANs}
SDYN-GANs provide data-driven methods for learning generative 
models for representing the $n^{th}$-order dynamics of stochastic systems.
As discussed in Section~\ref{sec_sdyn_gans}, we use adversarial 
learning losses $\ell^*(\theta_G)$ based on Integral Probability Metrics (IPMs)
with $\mathcal{F} = \{f \in \mathcal{H} \;|\; \|f\|_{\mathcal{H}} \leq 1\}$
corresponding to Generalized Moments Methods (GMMs)~\citep{Hansen1982} 
and Maximum Mean Discrepancy (MMD)~\citep{Gretton2012,Fortet1953,Smola2006}.  
Learning in SDYN-GANs only requires samples of the generator
and the training data, providing flexibility in the classes of 
generative models that can be used.  We develop approaches for dynamic generative 
models $G$ based on $m$-step stochastic discretization methods
for representing the trajectory samples.  This requires methods for 
computing during training the gradients of the second-order statistics 
in the loss for high dimensional samples.  We  
develop practical computational methods leveraging the structure of the loss 
$\hat{\ell}^*$, the probabilistic dependencies, and by developing 
adjoint approaches.  This provides efficient methods for computing 
the gradients needed for training SDYN-GANs.

\subsection{Generative Modeling based on $m$-Step Stochastic Integration Methods}
\label{sec_m_step_models}
We develop methods for classes of generative models $G$ based on stochastic
numerical integration methods~\citep{Platen1992,Jasuja2022,Atzberger2007}.  For
$n^{th}$-order stochastic dynamical systems we consider general stochastic
$m$-step numerical methods of the form
$\mb{X}_j =
\bsy{\Psi}_{j-1}(\mb{X}_{j-1},\ldots,\mb{X}_{j-m},\bsy{\omega}_{j-1};\mb{p}),$
with initial conditions $\mb{X}_0 =
\mb{x}_0(\mb{p}),\;\;\cdots\;\;\mb{X}_{m-1} = \mb{x}_{m-1}(\mb{p}).$
The $\mb{X}_j$ approximates $\mb{X}(t_j)$ when starting with the initial
conditions $\mb{X}_k = \mb{x}_k(\mb{p})$ at times $t_k$ with $k\in[0,m-1]$. The
$\bsy{\Psi}_{j-1}$ approximates the evolution of the stochastic dynamics from
time step $t_{j-1}$ to $t_j$, using information at times
$t_{j},\ldots,t_{j-m}$. Since the dynamics is stochastic, the numerical update
is stochastic and $\bsy{\Psi}_j(\cdot,\bsy{\omega};\mb{p})$ can be viewed as a
random variables on a probability space $(\Omega,\mathcal{B},\mu)$.  Here, for
the discretized system we use a collection of standard i.i.d. Gaussian samples
$\bsy{\omega} = \{\bsy{\omega}_{j}\}_j \in \Omega$. The $\mb{p}$ gives the
parameters of the numerical method, which can include the time-step
$\Delta{t}$, initial conditions, and other properties of the underlying
stochastic system being approximated. The above formulation includes implicit
methods where $\Psi$ would represent the update map from solving of the
implicit equations~\citep{Platen1992}.  In terms of the generative model $G$,
the samples are generated as $\mb{X}^{[i]} = G(\mb{Z}^{[i]};\theta_G)$, where
we take $\theta_G = \mb{p}$ and sample $\mb{Z}^{[i]} =
\{\bsy{\omega}_j^{[i]}\}_j$. 

For approximating solutions of stochastic and ordinary differential equations,
it is well-known the updates of numerical methods must be chosen with care to
ensure convergence. A common strategy to  obtain convergence is to impose a
combination of local accuracy in the approximation (consistency) along with
conditions more globally on how errors can accumulate
(stability)~\citep{Iserles2009,Burden2010,Platen1992}.  We develop our
generative model classes to achieve related properties by basing our learned
models $\bsy{\Psi}$ on stable and accurate stochastic integration methods. 

Methods used widely for first-order stochastic dynamical systems include
the $(m=1)$-step Euler-Maruyuma Methods~\citep{Maruyama1955}, Milstien
Methods~\citep{Milstein1975}, and Runge-Kutta Methods~\citep{Platen1992}.  For
higher-order accuracy or for $(n > 1)$-order dynamics, $m$-step multistep methods can
be developed, such as stochastic variants of Adams-Bashforth and
Adams-Moulten~\citep{Platen1992,Buckwar2006,Ewald2005}.  For physical systems
further considerations also lead to imposing additional requirements or 
alternative methods 
such as
approximate
time-reversibility~\citep{Verlet1967,Farago_Langevin_Num_Method_2013},
fluctuation-dissipation balance~\citep{Atzberger2007,AtzbergerTabak2015}, 
geometric and topological conditions~\citep{Lopez2022,Atzberger_GMLS_Surf_PDE_2019},
or
using exponential time-step integration~\citep{Jasuja2022,Atzberger2007}.
We develop methods for learning over generative model classes 
which can leverage the considerations of stability, accuracy, 
physical attributes, and other properties of the considered
stochastic systems.

\subsection{Gradient Methods for $m$-Step Generative Models}
\label{sec_m_step_gradients}
The training of the $m$-step generative models requires 
methods for computing the gradients of the loss 
$\nabla_{\theta_G} \hat{\ell}^*$ of equation~\ref{equ_tilde_ell_star}.
This poses challenges given the 
contributions of 
$\nabla_{\theta_G} \mb{X}^{[i]}(;\theta_G)$ where
$\theta_G = \mb{p}$, as discussed in Section~\ref{sec_sdyn_gans}.
Training requires computing contributions of the gradients of 
the discretized sample paths
$\{\mb{X}(t;\mb{p})\}_{t \in [0,\mathcal{T}]}$.
For this purpose, we will utilize the Implicit Function Theorem to develop
a set of adjoint methods for our $m$-step generative 
models~\citep{Jittorntrum1978,Cea1986,Strang2007,Smith2012}.  
Let $\mb{f} = \mb{f}(\mb{X}_0,\ldots,\mb{X}_N;\mb{p})$ be a
vector-valued function with components
\begin{eqnarray}
[\mb{f}]_{(j,d)}^{[i]} = \left[\mb{X}_j^{[i]} - 
\bsy{\Psi}_{j-1}(\mb{X}_{j-1}^{[i]},\ldots,\mb{X}_{j-m}^{[i]},\bsy{\omega}_{j-1};\mb{p})\right]^{(d)} = 0,
\end{eqnarray}
where $\mb{X}_k^{[i]} = \mb{X}^{[i]}(t_k)$ is the $i^{th}$ sample.  The
superscript index $d$ denotes the dimension component for $\mb{X}_k \in
\mathbb{R}^n$. To simplify the notation let $\mb{x} = \{\mb{x}^{[i]}\}_{i=1}^M
= \{(\mb{X}_0,\ldots,\mb{X}_{N_{\subtxt{$\mathcal{T}$}}})^{[i]}\}_{i=1}^M$, where there are $M$ samples
indexed by $i$. Denote by $g(\mb{x},p)=\hat{\ell}^*$ the loss function for which we need to
compute gradients in $p$ for training.  For each realization of $\bsy{\omega}^{[i]}$,
the trajectory samples are determined implicitly by $\mb{f}^{[i]}(\mb{x};p) =
0$.

The total derivative of $g$ is given by the chain-rule
\begin{eqnarray}
[\nabla_p g]_j = \frac{dg}{dp_j} = \frac{\partial g}{\partial p_j}
+ \sum_{k} \frac{\partial g}{\partial x_k} \frac{\partial x_k}{\partial p_j} = 
[\mb{g}_p + \mb{g}_x \mb{x}_p]_j.
\end{eqnarray}
We collect derivatives into vectors $\mb{g}_x \in \mathbb{R}^{1 \times
n_x},\mb{g}_p \in \mathbb{R}^{1 \times n_p} ,\mb{x}_p \in \mathbb{R}^{n_x
\times n_p}$, where $\mb{x} \in \mathbb{R}^{n_x}, \mb{p} \in \mathbb{R}^{n_p}$.

The derivatives of the loss $g$ in $\mb{x}$ and $\mb{p}$ are typically easier
to compute relative to the derivatives in $\mb{p}$ for $\mb{x}_p = \nabla_p
\mb{x}(p)$.  We can obtain expressions for the derivatives $\mb{x}_{\mb{p}}$
using the Implicit Function Theorem for $\mb{f}^{[i]}(\mb{x},\mb{p}) = 0$ with
$\mb{f}^{[i]} \in \mathbb{R}^{n_f^{[i]}}$ and under the assumption
$\mb{f}^{[i]}_{\mb{x}}$ is invertible~\citep{Jittorntrum1978,Smith2012}.  This gives
\begin{eqnarray}
\frac{df}{dp} = 
\mb{f}_x \mb{x}_p + \mb{f}_p = 0 \; \; \Rightarrow \; \; \mb{x}_p = -\mb{f}_x^{-1} \mb{f}_p = -J^{-1}\mb{f}_p. 
\end{eqnarray}
We assume throughout $n_f = n_x$, but distinguish these in the notation to make 
more clear the roles of the conditions.  The Jacobian in $x$ is given by $J = \mb{f}_x =
\nabla_x \mb{f} \in \mathbb{R}^{n_f \times n_x}$ for the map $\mb{y} =
\mb{f}(\mb{x};\mb{p})$ defined by fixed $p$ for mapping $\mb{x} \in
\mathbb{R}^{n_x}$ to $\mb{y} \in \mathbb{R}^{n_f}$. The $\mb{x}_p = \nabla_p
\mb{x}(p) \in \mathbb{R}^{n_x \times n_p}$ and $\mb{f}_p \in \mathbb{R}^{n_f
\times n_p}$. We can substitute this above to obtain 
\begin{eqnarray}
\nabla_p g = \mb{g}_p + \mb{g}_x(-\mb{f}_x^{-1}\mb{f}_p)
= \mb{g}_p + \left(-\mb{f}_x^{-T} \mb{g}_x^T\right)^T\mb{f}_p
=
\mb{g}_p - \left(J^{-T}\mb{g}_x^T\right)^T\mb{f}_p
= 
\mb{g}_p - \mb{r}^T\mb{f}_p.
\end{eqnarray}
We let $\mb{r} = J^{-T}\mb{g}_x^T \in \mathbb{R}^{n_f \times 1}$. To compute
the gradient $\nabla_p g$ we proceed in the following steps: (i) compute the
derivatives $\mb{g}_p, \mb{g}_x, \mb{f}_p$, (ii) solve the linear system
$J^T\mb{r} = \mb{g}_x^T$, (iii) evaluate the gradient using $\nabla_p g =
\mb{g}_p - \mb{r}^T \mb{f}_p$.

This provides methods for computing the gradients of $g$ using the solutions 
$\mb{r} \in \mathbb{R}^{n_f \times 1}$, which in general for the trajectory samples
will be more efficient than solving directly for 
$\mb{x}_p \in \mathbb{R}^{n_x\times n_p}$.  We emphasize that the  
construction of a dense $n_x\times n_x$ Jacobian and its inversion 
would be prohibitive, so instead we will solve the linear equations 
by approaches 
that use sparsity and only require computing the action of the 
adjoint $J^{T}$.  We show how this can be done for our $m$-step 
methods below and give more technical details 
in Appendix~\ref{appendix_m_step_recur}.

Our adjoint approaches are used to compute the gradient of the losses 
based on expectations of the following form and their gradients
\begin{eqnarray}
g^*(\mb{X}(\cdot),p) = \mathbb{E}_{\omega}[\phi(\mb{X}(\omega;p);p)] = \int \phi(\mb{X}(\omega;p);p) d\mu_\omega,
\;\;\;
\frac{dg^*}{dp} = \mathbb{E}[\nabla_p \phi] = \mathbb{E}[\nabla_p \tilde{g}].
\end{eqnarray}
We emphasize $g^*$ has all of its parameter
dependence explicit in the random variable $\mb{X}(\omega,p)$ and $\tilde{g} = \phi$, so the reference measure
$\mu_\omega$ has no $p$-dependence.  In practice, the expectations will be approximated 
by sample averages $\bar{g} = \frac{1}{M}
\sum_{i=1}^M \tilde{g}^{[i]}$, where $\tilde{g}^{[i]}$ denotes the $i^{th}$ sample.
In this case our adjoint approaches compute the gradients using 
\begin{eqnarray}
\nabla_p \bar{g}(\mb{x},\mb{p}) = 
\frac{1}{M} \sum_{i=1}^M \nabla_p \tilde{g}^{[i]},\;\;
\nabla_p \tilde{g}^{[i]} = 
\mb{\tilde{g}}_p^{[i]} - \mb{r}^{[i],T}\mb{f}_p^{[i]},\;\;
\mb{r}^{[i]} = J^{[i],-T}\mb{\tilde{g}}_x^{[i],T},
\end{eqnarray}
where $J^{[i],T} = \mb{f}_x^{[i],T}$.  We provide efficient direct methods for solving
these linear systems for $\mb{r}^{[i]}$ for the $m$-step methods 
in Appendix~\ref{appendix_m_step_recur}.

\subsection{Gradient Methods for the Second-Order Statistics in the Loss $\ell^*$}
\label{sec_backprop_loss}

The loss function $\tilde{\ell}^*$ in equation~\ref{equ_tilde_ell_star}
involves computing second-order statistics of the samples.  This is used to 
compare the samples $\mb{X}^{[i]} \sim \tilde{\mu}_G$ of the candidate
generative model with samples $\mb{Y}^{[i]}
\sim \tilde{\mu}_\mathcal{D}$ of the training data.  In computing gradients, this poses challenges
in using directly backpropagation on the forward computations, which can result in prohibitively 
large computational call graphs.  

We develop alternative methods utilizing special structure of the loss function $\tilde{\ell}^*$ 
to obtain more efficient methods for estimating the needed gradients for training.
From equation~\ref{equ_tilde_ell_star}, we see $\tilde{\ell}^*$ involves 
kernel calculations involving terms of the 
form $k(\mb{W}_1^{[i]},\mb{W}_2^{[j]};p)$, where $\mb{W}_1,\mb{W}_2 \in \{\mb{X},\mb{Y}\}$.  To obtain the gradient 
$\nabla_p \tilde{\ell}^*$, we need to compute the contributions of three terms, each of which
have the general form
\begin{eqnarray}
\label{equ_k_deriv}
\frac{d}{dp} k(\mb{W}_1^{[i]},\mb{W}_2^{[j]};p) 
&=& 
\partial_1 k\cdot \mb{W}_{1,p}^{[i]}
+ \partial_2 k\cdot \mb{W}_{2,p}^{[i]}
+ \partial_3 k. 
\end{eqnarray}
The $\partial_q k$ denote taking the partials in the $q^{th}$ input to
$k(\cdot,\cdot;p)$. The most challenging terms to compute are the partials in
the trajectory samples $\mb{W}_{a,p} = \partial \mb{W}_a/\partial p$ when
$\mb{W}_a=\mb{X}$.  In this case, we have $\mb{W}_{a,p} = \mb{X}_p$.  For $\mb{Y}$, 
we have $\mb{Y}_{p} = 0$, since the training data does not
depend on $p$.  We use that the gradients ultimately contract against the
vector-valued terms $\partial_1 k,$ $\partial_2 k$ in equation~\ref{equ_k_deriv}.  
As a consequence, we can
compute the terms by using $h(\mb{X}^{[i]}) = \mb{c}\cdot \mb{X}^{[i]}$.  This
is equivalent to computing derivatives when $\mb{c}$ is
treated as a constant, which yields $\nabla_p h(\mb{X}^{[i]}) = \mb{c}\cdot \mb{X}^{[i]}_p$.  
In other words, in the calculation after
differentiation, this term is then filled in with the numerical values $\mb{c}
= \partial_1 k$ or $\mb{c}=\partial_2 k$ for the final evaluation.  This is
useful, since this allows us to compute efficiently $\nabla_p
h(\mb{X}^{[i]})$ using our adjoint methods in
Section~\ref{sec_m_step_gradients}.  To compute the partial $\partial_3 k =
\partial k/\partial p$ we can use backpropagation while holding the other
inputs fixed.

To compute the contributions to the gradient $\tilde{\ell}^*$, we consider the
first term $k(\mb{X}^{[i]},\mb{X}^{[j]};p)$ in equation~\ref{equ_tilde_ell_star} and
treat the sums over the samples using
\begin{eqnarray}
\frac{d}{d{p}} \sum_{i\neq j} k(\mb{X}^{[i]},\mb{X}^{[j]};p) 
&=& \sum_{i_0} \mb{c}_1(i_0)^T \mb{X}_p^{[i_0]} + 
\sum_{j_0} \mb{c}_2(j_0)^T \mb{X}_p^{[j_0]} + \mb{c}_3,
\end{eqnarray}
where 
\begin{eqnarray}
\mb{c}_1(i_0) = \sum_{j \neq i_0} \partial_1 k(\mb{X}^{[i_0]},\mb{X}^{[j]};p),\;\;\;  
\mb{c}_2(j_0) = \sum_{i \neq j_0} \partial_2 k(\mb{X}^{[i]},\mb{X}^{[j_0]};p),\;\;\; 
\mb{c}_3 = \sum_{i \neq j} \partial_3 k(\mb{X}^{[i]},\mb{X}^{[j]};p).
\end{eqnarray}
To compute the contributions of the terms $\mb{X}_p^{[i_0]} = \frac{d}{dp}\mb{X}^{[i_0]}$ and 
$\mb{X}_p^{[j_0]} = \frac{d}{dp}\mb{X}^{[j_0]}$, we use the adjoint methods discussed in
Section~\ref{sec_m_step_gradients} and Appendix~\ref{appendix_m_step_recur}.
For the second term contributing to $\tilde{\ell}^*$ in equation~\ref{equ_tilde_ell_star}, we use
\begin{eqnarray}
\frac{d}{d{p}} \sum_{i,j} k(\mb{X}^{[i]},\mb{Y}^{[j]};p) 
= \sum_{i_0} \mb{\tilde{c}}_1(i_0)^T \mb{X}_p^{[i_0]} + \mb{\tilde{c}}_3,
\end{eqnarray}
\begin{eqnarray}
\mb{\tilde{c}}_1(i_0) = \sum_{j} \partial_1 k(\mb{X}^{[i_0]},\mb{Y}^{[j]};p),\;\;
\mb{\tilde{c}}_3 = \sum_{i,j} \partial_3 k(\mb{X}^{[i]},\mb{X}^{[j]};p).
\end{eqnarray}
For the third term in equation~\ref{equ_tilde_ell_star}, the only dependence on $p$
is through the direct contributions to the kernel, since $\mb{Y}^{[i]}$ does not 
depend on $p$.  We 
compute the gradient using 
\begin{eqnarray}
\frac{d}{d{p}} \sum_{i\neq j} k(\mb{Y}^{[i]},\mb{Y}^{[j]};p) 
= \mb{\hat{c}}_3,
\;\; \mb{\hat{c}}_3 = \sum_{i \neq j} \partial_3 k(\mb{Y}^{[i]},\mb{Y}^{[j]};p).
\end{eqnarray}
For each term, the $\partial_j k = \partial_j k(w_1,w_2;p)$ can be computed using
evaluations of $k$ and local backpropagation methods.  We use this to  
compute $\mb{c}_j, \mb{\tilde{c}}_j, \mb{\hat{c}}_j$.
While the full gradient sums over all the samples, in practice we use
stochastic gradient descent. This allows for using smaller mini-batches of samples 
to further speed up calculations.  For SDYN-GANs, we use these approaches
to implement custom backpropagation methods for gradient-based training of 
our $m$-step generative models.  We also provide further details in 
Appendix~\ref{appendix_m_step_recur}.

\section{Results}
\label{sec_results}
We present results for SDYN-GANs in learning generative models for second-order stochastic dynamics.
This is motivated by the physics
of microscale mechanical systems and their inertial stochastic dynamics.  We
develop methods for learning force-laws, damping coefficients, and
noise-related parameters.  We present results for learning physical models for
a micro-mechanical system corresponding to an Inertial Ornstein-Uhlenbeck
(I-OU) Process in Section~\ref{sec_OU_process}.  We then consider learning
non-linear force-laws for a particle system having inertial Langevin Dynamics
in Section~\ref{sec_results_force_laws}.  

We investigate in these studies the role of the time-scale $\tau$ for the
trajectory sampling and the different training protocols based on sampling from
the distributions of the (i) full-trajectory, (ii) conditionals, and (iii)
marginals.  
The results show a few strategies for training SDYN-GANs to obtain representations of stochastic systems.

\subsection{Discretization of Inertial Second-order Stochastic Dynamics}
\label{sec_discrete_stoch_dyn}
We consider throughout second-order inertial stochastic dynamics of the form 
\begin{eqnarray}
md\mb{V}(t) = -\gamma\mb{V}(t)dt + \mb{F}(\mb{X}(t))dt 
+ \sigma d\mb{W}(t), \;\;\; d\mb{X}(t) = \mb{V}(t)dt.
\label{equ_stoch_dyn_inertial}
\end{eqnarray}
For SDYN-GANs, our generative models use the following $(m=2)$-step stochastic
numerical integrator~\citep{Farago_Langevin_Num_Method_2013}.  The integrator
can be expressed as a Velocity-Verlet-like scheme~\citep{Verlet1967} of the
form 
\begin{eqnarray}
\nonumber
\mb{X}_{n+1} & = & \mb{X}_n + b\Delta{t}\mb{V}_n 
+ \frac{b\Delta{t}^2}{2m}\mb{F}_n 
+ \frac{b\Delta{t}}
{2m}\bsy{\eta}_{n+1}, \;\;\;\; \;\;\;\;
a = \left({1 - \frac{\gamma \Delta{t}}{2m}}\right)
 \left({1 + \frac{\gamma \Delta{t}}{2m}}\right)^{-1},
\\
\mb{V}_{n+1} & = & a\mb{V}_n + \frac{\Delta{t}}{2m}
\left(a \mb{F}_n + \mb{F}_{n+1}\right) 
+ \frac{b}{m}\bsy{\eta}_{n+1}, \;\; \;\;\;\;\;\;
 b = \left({1 + \frac{\gamma \Delta{t}}{2m}}\right)^{-1}.
\label{equ_farago_method}
\end{eqnarray}
We refer to this discretization of the stochastic dynamics as the Farago
Method, which has been shown to have good stability and statistical properties
in~\citep{Farago_Langevin_Num_Method_2013}.  The thermal forcing is
approximated by Gaussian noise $\bsy{\eta}_n$ with mean zero and variance
$\langle \bsy{\eta}_n \bsy{\eta}_{\ell}^T \rangle = \sigma^2 I
\delta_{n,\ell}\Delta{t}$ for time-step $\Delta{t}$. This can be expressed as
the 2-stage multi-step method as $\mb{X}_{n+1} =
\bsy{\Psi}_n(\mb{X}_n,\mb{X}_{n-1},\bsy{\omega}_n;\mb{p})$ with
$\bsy{\Psi}_n = 
2b\mb{X}_n - a\mb{X}_{n-1} + \frac{b\Delta{t}^2}{m}\mb{F}_n +
\frac{b\Delta{t}}{2m}\left(\bsy{\eta}_{n+1} + \bsy{\eta}_n\right).
$
Our generative models use throughout this stochastic numerical discretization
for approximating the inertial second-order stochastic dynamics of
equation~\ref{equ_stoch_dyn_inertial}.

\subsection{Learning Physical Models by Generative Modeling: Inertial Ornstein-Uhlenbeck Processes} 
\label{sec_OU_process} 
Consider the Ornstein-Uhlenbeck (I-OU) process~\citep{Uhlenbeck1930} which have the 
stochastic dynamics
\begin{eqnarray}
md\mb{V}(t) = -\gamma\mb{V}(t)dt - K_0\mb{X}dt + \mb{F}_0dt 
+ \sigma d\mb{W}(t), \;\;\; d\mb{X}(t) = \mb{V}(t)dt.
\label{equ_ou_process}
\end{eqnarray}
The $\mb{X},\mb{V} \in \mathbb{R}^n$.  Related dynamics arise when studying
molecular systems~\citep{Uhlenbeck1930,Reichl1997,Frenkel2001,Yushchenko2012},
microscopic devices~\citep{GomezSolano2010,Goerlich2021,Dudko2002}, problems in
finance and economics~\citep{Steele2001,Vasicek1977,Schoebel1999,AitSahalia2007}, 
and many other applications~\citep{Large2021,Wilkinson2012,Ricciardi1979,Giorgini2020}. 

From the perspective of mechanics, these dynamics can be thought of as modeling
a microscopic bead-spring system within a viscous solvent fluid. In this case,
$\mb{X}$ is the position and $\mb{V}$ is the velocity with $\mb{X},\mb{V} \in
\mathbb{R}^3$.  The $m$ is the bead's mass, $\gamma$ is the level of viscous
damping, $\sigma = \sqrt{2k_B{T}\gamma}$ gives the strength of the fluctuations
for temperature $T$, with $k_B$ denoting the Boltzmann
factor~\citep{Reichl1997}.  The conservative forces include a harmonic spring
with stiffness $K_0$ and a constant force $\mb{F}_0 = -mg \mb{e}_z$, such as
gravity with coefficient $g$ and $\mb{e}_z = (0,0,1)$.  This gives the
potential energy $U(\mb{X}) = \frac{1}{2}K_0\mb{X}^2 - \mb{F}_0\cdot\mb{X}$
yielding the  conservative forces as $\mb{F}(\mb{X}) = -\nabla_{\mb{X}} U$.  We
use for our generative models the dynamics of equation~\ref{equ_ou_process}
approximated by the $(m=2)$-step stochastic numerical discretization discussed
in Section~\ref{sec_discrete_stoch_dyn}.  We denote by $\theta$ the collection
of parameters to be adjusted as $\theta = (K_0,\gamma,k_BT)$. 

\begin{figure}[h]
\centerline{\includegraphics[width=0.99\columnwidth]{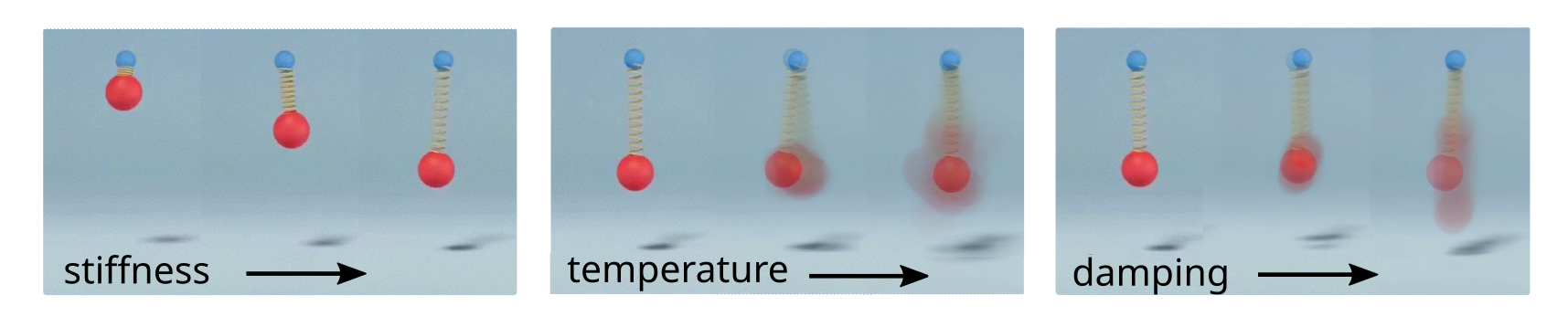}}
\caption{\textbf{Bead-Spring Mechanics: Inertial Ornstein-Uhlenbeck Process.}  We consider the 
stochastic dynamics of the three dimensional I-OU process under different
physical conditions including when (i) stiffness decreases \textit{(left)},
(ii) temperature increases \textit{(middle)}, or (iii) damping decreases
\textit{(right)}.}
\label{fig_model_3d}
\end{figure}

\begin{figure}[h]
\centerline{\includegraphics[width=0.99\columnwidth]{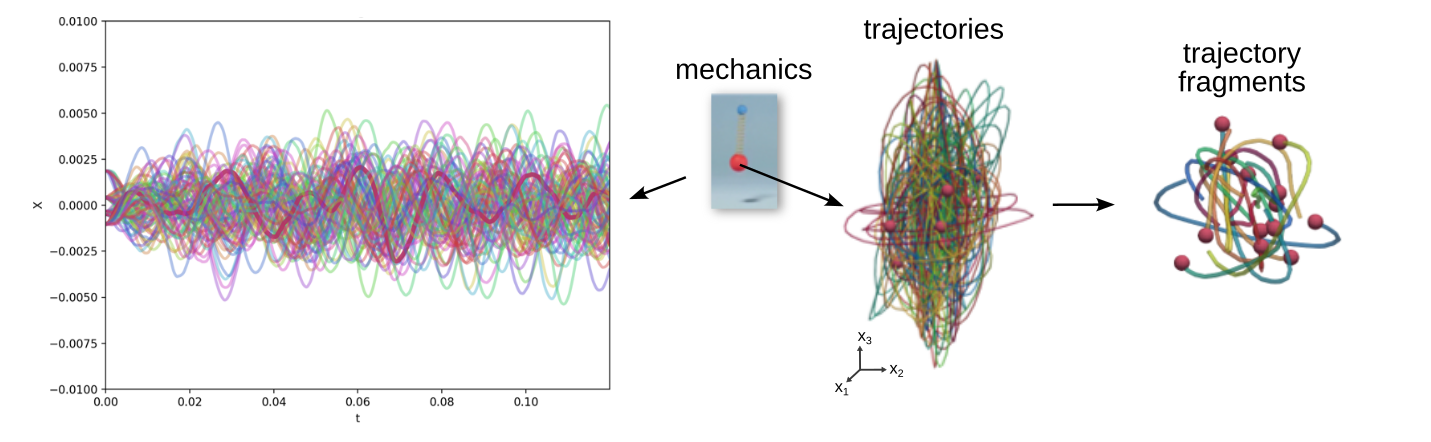}}
\caption{\textbf{Trajectory Training Data: Inertial Ornstein-Uhlenbeck Process.} 
We show samples of the trajectory data used for training models.  The z-component of 
the trajectory data is shown on the \textit{(left)}.  The bead-spring system
and traces of the trajectory in three-dimensional space is shown in the \textit{(middle)}. 
The fragments of the trajectory used during the different training protocols 
discussed in Section~\ref{sec_training_protocols}, on the \textit{(right)}.
}
\label{fig_traj_3d}
\end{figure}

The data-driven task is to learn a generative stochastic model
$G(\cdot;\theta_G)$ from observations of trajectories of the system
configurations
$\mb{X}_{[0,\mathcal{T}]} = \{\mb{X}(t)\}_{t \in[0,\mathcal{T}]}$,
so that $G(\mb{Z};\theta_G) \sim \mb{X}_{[0,\mathcal{T}]}$.  The 
$\mb{Z}$ is the noise source, here taken to be a high dimensional 
vector-valued Gaussian with i.i.d components. 
The
$\mb{V}(t)$ is not observed directly.  We show samples of the training data 
in Figure~\ref{fig_traj_3d}.  Training uses discretized 
trajectories and trajectory fragments of the form
$(\mb{X}(t_1),\ldots,\mb{X}(t_m))$.
Without using further structure of the
data, this task can be challenging given the high
dimensional distribution of trajectories $\mb{X}(t)$.  

The SDYN-GANs uses the probabilistic dependence within the trajectories and the
adversarial learning approaches discussed in Section~\ref{sec_sdyn_gans}.  This
provides flexible methods for learning general models for a wide variety of
tasks, even when the generative models may not have readily expressible
probability distributions.  By using MMD within our SDYN-GANs methods allows
for further inductive biases to be introduced based on the type of kernel
employed.  This can be used to emphasize key features of the empirical
distributions being compared. For the OU process studies, we use a RKHS with
Rational-Quadratic Kernel
(RQK)~\citep{Rasmussen2006,Duvenaud_Kernel_Cookbook_2014} given by 
$k(x,y) = \left(1 + \frac{\|x - y\|^2}{2\alpha\ell^2}\right)^{-\alpha}.$ 
The $\alpha$ determines the rate of decay and $\ell$ provides a characteristic
scale for comparisons.  The RQK is a characteristic kernel allowing for general
comparison of probability distributions~\citep{Sriperumbudur2010}.  It has the 
advantage for gradient-based learning of not having an exponentially decaying 
tail yielding better behaviors in 
high dimensions~\citep{Rasmussen2006,Duvenaud_Kernel_Cookbook_2014}. 

\begin{figure}[h]
\centerline{\includegraphics[width=0.99\columnwidth]{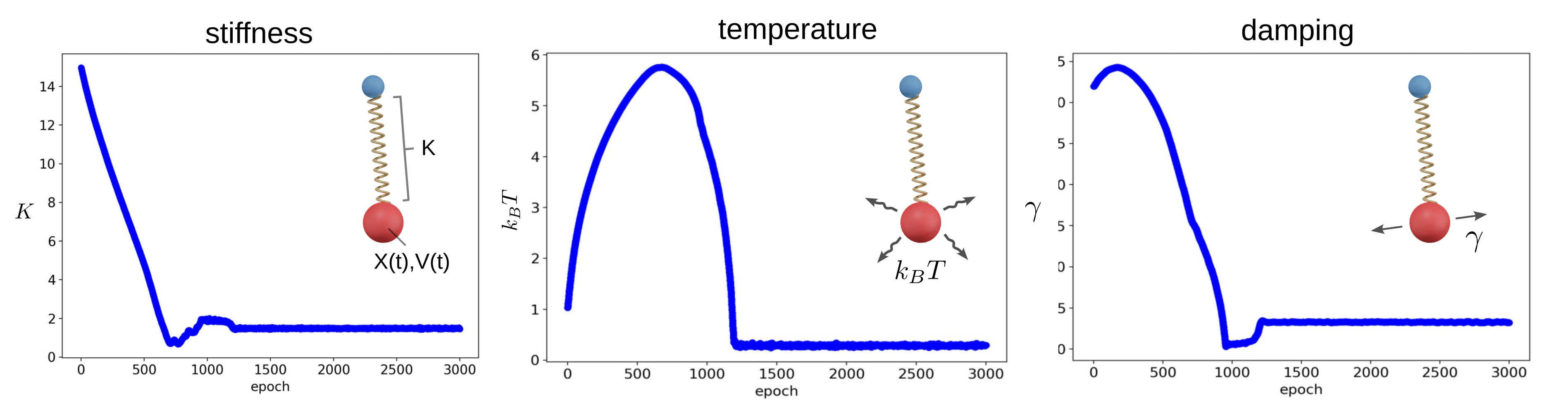}}
\caption{\textbf{Inertial Ornstein-Uhlenbeck Process: Learning the Stochastic Dynamics.}
We use SDYN-GANs to learn generative models for the observed trajectories of
the stochastic dynamics.  We learn simultaneously the drift
and diffusive contributions to the dynamics.  The target values 
for this trial were $K_0 = 1.5$, $\gamma = 3.2$ and $k_B{T} = 0.1$.
We see early in training the
stochastic parameters become large to minimize the loss by 
enhancing the overlap of the samples of the candidate generative 
model with the observed samples.  As the stiffness and damping 
parameters become more accurate during training, the stochastic parameters
decrease toward the target values.
}
\label{fig_OU_learning}
\end{figure}

We show a typical training behavior for learning simultaneously parameters
$\theta = (K_0,\gamma,k_BT)$  in Figure~\ref{fig_OU_learning}. We find that
training initially proceeds by increasing $k_BT$ which broadens the generative
model distribution to overlap the training data samples to reduce the loss.
However, we see that once the stiffness $K_0$ and damping $\gamma$
parameters approach reasonable values, around the epoch $\sim 750$, the
temperature $k_BT$ rapidly decreases to yield models that can more accurately
capture the remaining features of the observation data. We see for this
training trial that all parameters are close to the correct values around epoch
$\sim 1,200$, see Figure~\ref{fig_OU_learning}.

To investigate further the performance of SDYN-GANs, we perform  training using
a few different approaches.  This includes varying the time-scale $\tau$
overwhich we sample the system configurations $\mb{X}(t)$ and the type of
distributions implicitly probed.  For instance, we use the full-trajectory over 
a time-duration $\mathcal{T}$, while considering conditionals or marginals 
statistics over shorter time durations.  For this purpose, we use the different training   
protocols and time-scales $\tau$ discussed in Section~\ref{sec_training_protocols}.  
Throughout, we use the default target model parameters $m=0.1,K_0=1.5,\gamma=3.2,k_BT=0.1$ and
kernel parameters $\alpha=2,\ell=0.01$.
For the different SDYN-GANs methods and training protocols,
we show the accuracy of our learned generative models in the 
Tables~\ref{table_OU_accuracy_K_0}--\ref{table_OU_accuracy_gamma}.  

We find the stiffness and damping mechanics each can be learned with an accuracy of around 
$5\%$ or less.  The thermal fluctuations of the system were more challenging to learn
from the trajectory observations.  This appeared to arise from other contributing factors
to perceived fluctuations associated with sampling errors and a tendency to over-estimate
the thermal parameters to help enhance alignment of the candidate generative model with
the trajectory data.  We found accuracies of only around $50\%$ for the thermal 
fluctuations of the system. We found the time delay $\tau$ used in probing the temporal
sub-sampling of the dynamics can play a significant role in the performance of the 
learning methods.

\begin{table}[ht]
\centering
{
\fontsize{9}{12}\selectfont

\begin{tabular}{l|l|l|l}
\hline 
\rowcolor{atz_table2}
\multicolumn{4}{l}{\textbf{Accuracy $K_0$ vs $\tau$-Delay}} \\
\hline
\rowcolor{atz_table1}
\textbf{$\tau$-Delay} & \textbf{1.70e-02} & \textbf{1.19e-02} & \textbf{5.83e-03} \\
\hline
MMD-full-traj & $1.03e\sminus 02 \pm 1.1e\sminus 02$ & $2.62e\sminus 02 \pm 2.4e\sminus 02$ & $1.13e\sminus 02 \pm 7.8e\sminus 03$ \\
\hline
MMD-conditionals & $9.19e\sminus 03 \pm 1.9e\sminus 03$ & $2.22e\sminus 02 \pm 1.7e\sminus 02$ & $2.66e\sminus 02 \pm 4.5e\sminus 02$ \\ 
\hline
MMD-marginals & $1.54e\sminus 02 \pm 6.7e\sminus 03$ & $8.92e\sminus 03 \pm 1.1e\sminus 02$ & $4.37e\sminus 02 \pm 4.5e\sminus 02$  \\
\hline
\rowcolor{atz_table1}
\textbf{$\tau$-Delay} &\textbf{4.08e-03} & \textbf{2.86e-03} & \textbf{2.00e-03} \\
\hline
MMD-full-traj & $1.62e\sminus 02 \pm 1.5e\sminus 02$ & $8.02e\sminus 03 \pm 5.2e\sminus 03$ & $1.37e\sminus 02 \pm 7.0e\sminus 03$ \\
\hline
MMD-conditionals & $1.03e\sminus 01 \pm 1.9e\sminus 01$ & $8.52e\sminus 01 \pm 1.7e\splus00$ & $5.68e\sminus 01 \pm 1.1e\splus00$ \\
\hline
MMD-marginals & $4.44e\sminus 02 \pm 8.0e\sminus 02$ & $1.63e\splus00 \pm 3.2e\splus00$ & $1.72e\splus00 \pm 3.4e\splus00$ \\
\hline
\end{tabular}
}
\caption{\textbf{Accuracy $K_0$ verses $\tau$-Delay: Inertial Ornstein-Uhlenbeck Process.} 
SDYN-GANs training performed using the protocols over the (i) full trajectory, 
(ii) marginals, and (iii) conditionals, as discussed in Section~\ref{sec_training_protocols}. 
The accuracy of the stiffness estimate $\tilde{\phi}$ for $\phi = K_0$ 
given by the relative error
$\epsilon_\subtxt{rel} = |\tilde{\phi} - \phi|/|\phi|$.  
The number of epochs was $3000$.  Standard deviations are reported for 
$4$ runs. The total duration of all trajectories are
$t_n=1.8\times 10^{-2}$ over $n=18$ steps with $\Delta{t} = 10^{-3}$.
} 
\label{table_OU_accuracy_K_0}
\end{table}

For the temporal sampling, we found when the $\tau$ time-scales are relatively
large the models were learned more accurately.  This appears to be related to
the samples better probing the time-scales in the trajectory data overwhich the
stiffness and damping make stronger contributions.  The longer trajectory
durations appear to provide a better signal for learning these properties.  As
$\tau$ became smaller, we found the accuracy of the learned models decreased
for this task. 

We further found the training protocol chosen could have a significant impact
on the accuracy of the learned models.  For the current task, we found throughout 
that either the full-trajectory protocols or marginals protocols tended 
to perform the best.  A notable feature of the marginals protocol is that the trajectory
samples include the same fragments of the trajectory arising in the
conditionals, particularly those starting near the sampled initial conditions.  
This appears to have contributed to the marginals out-performing the conditionals,
the latter of which contain less information about the longer-time dynamics.  

\begin{table}[ht]
\centering
{
\fontsize{9}{12}\selectfont

\begin{tabular}{l|l|l|l}
\hline 
\rowcolor{atz_table2}
\multicolumn{4}{l}{\textbf{Accuracy $k_B{T}$ vs $\tau$-Delay}} \\
\hline
\rowcolor{atz_table1}
\textbf{$\tau$-Delay} & \textbf{1.70e-02} & \textbf{1.19e-02} & \textbf{5.83e-03} \\
\hline
MMD-full-traj & $5.36e\sminus 01 \pm 9.5e\sminus 01$ & $1.71e\splus00 \pm 3.4e\splus00$ & $3.62e\splus00 \pm 7.1e\splus00$ \\ 
\hline
MMD-conditionals & $6.43e\sminus 01 \pm 1.1e\splus00$ & $1.46e\splus00 \pm 2.6e\splus00$ & $1.92e\splus01 \pm 3.8e\splus01$ \\ 
\hline
MMD-marginals & $6.54e\sminus 01 \pm 1.2e\splus00$ & $1.51e\splus00 \pm 2.9e\splus00$ & $1.39e\splus01 \pm 2.7e\splus01$ \\ 
\hline
\rowcolor{atz_table1}
\textbf{$\tau$-Delay} &\textbf{4.08e-03} & \textbf{2.86e-03} & \textbf{2.00e-03} \\
\hline
MMD-full-traj & $3.47e\splus00 \pm 6.7e\splus00$ & $6.43e\splus00 \pm 1.3e\splus01$ & $6.90e\splus00 \pm 1.4e\splus01$ \\ 
\hline
MMD-conditionals & $2.42e\splus01 \pm 4.7e\splus01$ & $9.54e\splus01 \pm 1.7e\splus02$ & $9.38e\splus01 \pm 1.7e\splus02$ \\ 
\hline
MMD-marginals & $1.85e\splus01 \pm 3.7e\splus01$ & $3.95e\splus01 \pm 7.9e\splus01$ & $3.82e\splus01 \pm 7.6e\splus01$ \\ 
\hline
\end{tabular}
}
\caption{\textbf{Accuracy $k_B{T}$ verses $\tau$-Delay: Inertial Ornstein-Uhlenbeck Process.} 
SDYN-GANs training performed using the protocols over the (i) full trajectory, 
(ii) marginals, and (iii) conditionals, as discussed in Section~\ref{sec_training_protocols}. 
The accuracy of the temperature estimate $\tilde{\phi}$ for $\phi = k_B{T}$ 
given by the relative error
$\epsilon_\subtxt{rel} = |\tilde{\phi} - \phi|/|\phi|$.  
The number of epochs was $3000$.  Standard deviations are reported for 
$4$ runs. The total duration of all trajectories are
$t_n=1.8\times 10^{-2}$ over $n=18$ steps with $\Delta{t} = 10^{-3}$.
} 
\label{table_OU_accuracy_KBT}
\end{table}

We found overall a better job was
done by the full-trajectory training protocol which uses more steps along the trajectory 
than the marginals and conditionals
protocols.  The marginals also tended to do better than the conditionals.
As an exception, we did find in a few cases for small $\tau$ 
the conditionals did a better job than the marginals in estimating the 
stiffness.  This was perhaps since when working over shorter time durations,
the more temporally localized sampling near the same initial conditions in 
the conditionals provide more consistent and less obscured information on 
the stiffness responses, relative to the full set of trajectory fragments
contributing in the marginals protocols.

\begin{table}[ht]
\centering
{
\fontsize{9}{12}\selectfont

\begin{tabular}{l|l|l|l}
\hline 
\rowcolor{atz_table2}
\multicolumn{4}{l}{\textbf{Accuracy $\gamma$ vs $\tau$-Delay}} \\
\hline
\rowcolor{atz_table1}
\textbf{$\tau$-Delay} & \textbf{1.70e-02} & \textbf{1.19e-02} & \textbf{5.83e-03} \\
\hline
MMD-full-traj & $1.53e\sminus 02 \pm 1.5e\sminus 02$ & $3.31e\sminus 02 \pm 2.5e\sminus 02$ & $2.91e\sminus 02 \pm 4.3e\sminus 02$ \\ 
\hline
MMD-conditionals & $2.85e\sminus 02 \pm 2.3e\sminus 02$ & $6.57e\sminus 02 \pm 6.0e\sminus 02$ & $2.29e\sminus 02 \pm 3.3e\sminus 02$ \\ 
\hline
MMD-marginals & $1.67e\sminus 02 \pm 8.4e\sminus 03$ & $3.34e\sminus 02 \pm 1.9e\sminus 02$ & $6.20e\sminus 02 \pm 8.9e\sminus 02$ \\ 
\hline
\rowcolor{atz_table1}
\textbf{$\tau$-Delay} &\textbf{4.08e-03} & \textbf{2.86e-03} & \textbf{2.00e-03} \\
\hline
MMD-full-traj & $5.20e\sminus 02 \pm 4.7e\sminus 02$ & $4.70e\sminus 02 \pm 4.6e\sminus 02$ & $2.20e\sminus 02 \pm 1.8e\sminus 02$ \\ 
\hline
MMD-conditionals & $2.00e\sminus 01 \pm 3.7e\sminus 01$ & $1.91e\splus00 \pm 3.8e\splus00$ & $1.88e\splus00 \pm 3.7e\splus00$ \\ 
\hline
MMD-marginals & $1.90e\sminus 01 \pm 3.0e\sminus 01$ & $5.54e\splus00 \pm 1.1e\splus01$ & $5.47e\splus00 \pm 1.1e\splus01$ \\ 
\hline
\end{tabular}
}
\caption{\textbf{Accuracy $\gamma$ verses $\tau$-Delay: Inertial Ornstein-Uhlenbeck Process.} 
SDYN-GANs training performed using the protocols over the (i) full trajectory, 
(ii) marginals, and (iii) conditionals, as discussed in Section~\ref{sec_training_protocols}. 
The accuracy of the damping estimate $\tilde{\phi}$ for $\phi = \gamma$ 
given by the relative error
$\epsilon_\subtxt{rel} = |\tilde{\phi} - \phi|/|\phi|$.  
The number of epochs was $3000$.  Standard deviations are reported for 
$4$ runs. The total duration of all trajectories are
$t_n=1.8\times 10^{-2}$ over $n=18$ steps with $\Delta{t} = 10^{-3}$.
} 
\label{table_OU_accuracy_gamma}
\end{table}

For the longer $\tau$ cases when comparing with the conditionals, 
the marginals tended to more accurately estimate simultaneously 
the stiffness and damping.  This likely arose, since 
unlike the conditionals case, the marginals do not require the initial parts of 
the trajectory fragments being compared to share the same initial conditions.  
For the general short-duration fragments, there is additional information on 
the long-time dynamical responses contributing to the marginals.  In summary,
the results suggest using the full-trajectory protocols when possible and 
relatively long $\tau$ sampling.  When one is only able to probe short-duration
trajectory fragments, the results suggest the marginals provide
some advantages over the conditionals.  Which protocols work best 
may be dependent on the task and the time-scales associated
with the dynamical behaviors being modeled.  These results show a 
few strategies for using SDYN-GANs to learn generative models for 
stochastic systems.

\subsection{Learning Unknown Force-Laws with SDYN-GANs Generative Modeling}
\label{sec_results_force_laws}

\begin{figure}[ht]
\centerline{\includegraphics[width=0.99\columnwidth]{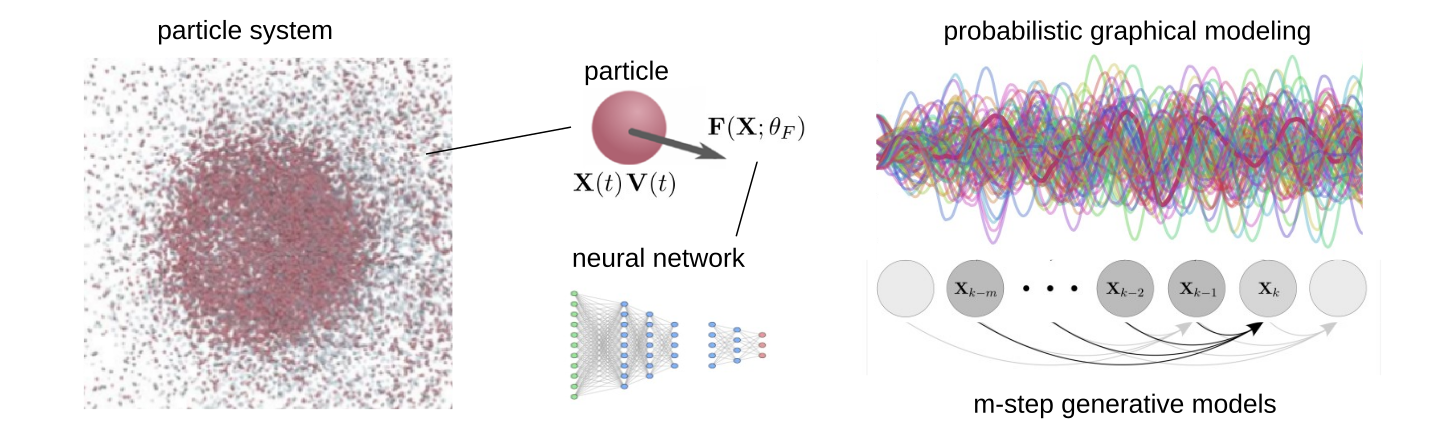}}
\caption{\textbf{Learning Unknown Force-Laws.}
SDYN-GANs is used to learn an unknown force-law from observation of particle
trajectories \textit{(left)}.  The learned force-law can be represented using
neural networks or other model classes \textit{(middle)}.  The generative
models use the dependence in the probabilities of the stochastic trajectories
using $m$-step stochastic numerical discretizations \textit{(right)}.  The
models are learned using a few different training protocols using statistics of
the (i) full-trajectory, (ii) conditionals, and (iii) marginals, as discussed
in Section~\ref{sec_training_protocols}. 
}
\label{fig_force_laws}
\end{figure}

\begin{figure}[h]
\centerline{\includegraphics[width=0.99\columnwidth]{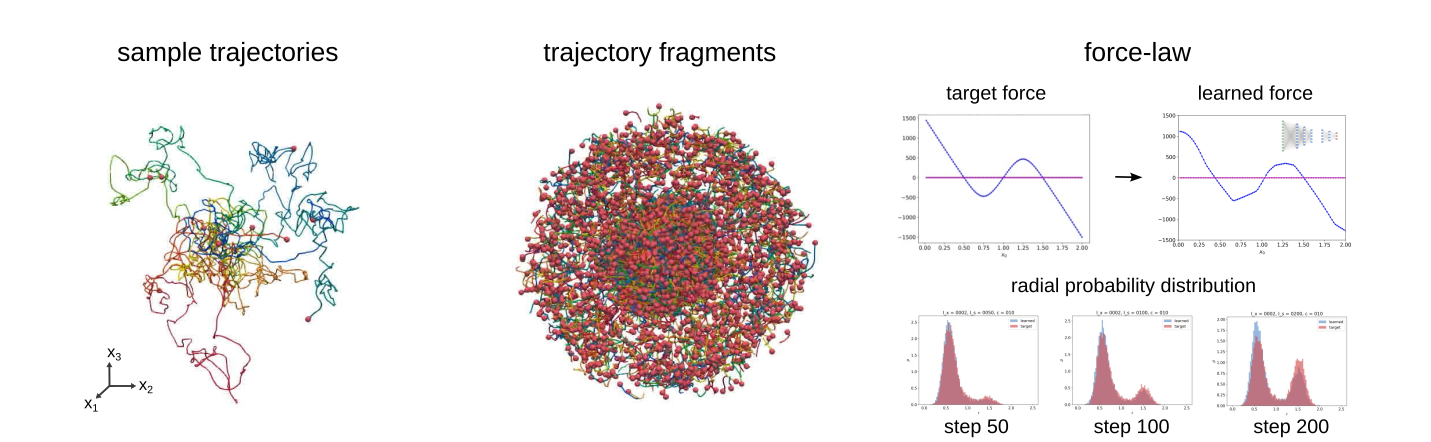}}
\caption{\textbf{Learning Unknown Force Laws: Trajectory Data.} 
We show a few samples of the particle trajectories \textit{(left)}.  The force-law 
is learned from samples of trajectory fragments using the different
training protocols discussed in Section~\ref{sec_training_protocols}, 
\textit{(middle)}.  SDYN-GANs is used to learn a neural network representation 
of the force-law from observations of trajectories of the particle system.
We evaluate the final accuracy of the 
learned model by considering the relative $L^1$-norm of the marginal 
time-dependent radial probability distributions of samples of the learned
model compared 
with the target particle system
\textit{(right)}.
}
\label{fig_force_law_traj}
\end{figure}

We show how SDYN-GANs can be used to learn
force-laws $\mb{F}(\mb{X})$ from observations of trajectories
of stochastic systems.  We consider the inertial Langevin 
dynamics of particle systems of the form 
\begin{eqnarray}
\label{equ_langevin_force_law}
md\mb{V}(t) = -\gamma\mb{V}(t)dt +\mb{F}(\mb{X};\theta_F)dt 
+ \sqrt{2k_B{T}\gamma}d\mb{W}(t), \;\;\; 
d\mb{X}(t) = \mb{V}(t)dt,
\end{eqnarray}
where
$\mb{X},\mb{V} \in \mathbb{R}^3$. The force-law $\mb{F}(\mb{X})$ will be
modeled using Deep Neural Networks (DNNs)~\citep{LeCun2015,Goodfellow2016}.
In this case, $\theta_F$ are the DNN weights and biases to be learned.  We
consider here the case of radial potentials $\mb{F}(\mb{X}) =
\mb{F}(\|\mb{X}\|) = F(r)\mb{e}_r$ with $\mb{e}_r = \mb{X}/\|\mb{X}\|$. For
simplicity, we focus on the case where $m,\gamma,T$ are fixed, but these also
could be learned in conjunction with the force-law. The data-driven task is to learn
DNN representations $\mb{F}(\mb{X};\theta_F)$ for the force-law from the configuration 
trajectories $\mb{X}_{[0,\mathcal{T}]} = \{\mb{X}(t)\}_{t \in[0,\mathcal{T}]}$ 
discretized as $(\mb{X}(t_1),\ldots,\mb{X}(t_n))$,
see Figure~\ref{fig_force_laws}.  The $\mb{V}(t)$ is not observed directly.
We use for our generative models the Langevin dynamics of
equation~\ref{equ_langevin_force_law} approximated by the $(m=2)$-step stochastic
numerical discretizations discussed in Section~\ref{sec_discrete_stoch_dyn}.  We
show samples of the training data in Figure~\ref{fig_force_law_traj}.

In our generative models, we use DNN architectures 
of multi-layer perceptrons with input-hidden-output sizes
\textit{$n_\subtxt{in}{\sdash}n_1 {\sdash}n_2{\sdash}n_3{\sdash}
n_\subtxt{out}$} and LeakyReLU~\citep{Xu_LeakyReLU_2015} units with
\textit{slope}=-0.01. In the DNNs we use $n_\subtxt{out} = n_\subtxt{in} = 1$,
with default values of $n_1=n_2=n_3=100$ and biases for all layers except for
the last layer. The DNNs serve to model the $F$ in the radial force $\mb{F} =
F(r;\theta_F)\mb{e}_r$.

We use SDYN-GANs to learn generative models by probing trajectories on different
time-scales $\tau$ and training protocols that are based on sampling the (i)
full-trajectory, (ii) conditionals, and (iii) marginals, as 
discussed in Section~\ref{sec_training_protocols}.  To train the models,
we simulate the trajectories of the candidate generative models and 
compare with the samples of the observation training data.  To
characterize the final accuracy of the learned models, 
we use the relative errors under the $L^1$-norm of the 
time-dependent radial probability distributions to compare the 
samples of the generative model with the target particle system.
We emphasize the $L^1$-error is not used during training, but only
on the final models to characterize the accuracy of the stochastic
trajectories they generate.  We train using SDYN-GANs
with different time-scales $\tau$ and training protocols,
with our results reported in Table~\ref{table_dw2_DNN_L1_err}.

\begin{table}[ht]
\centering
{
\fontsize{9}{12}\selectfont

\begin{tabular}{l|l|l|l}
\hline 
\rowcolor{atz_table2}
\multicolumn{4}{l}{\textbf{Accuracy vs $\tau$-Delay}} \\ 
\hline
\rowcolor{atz_table1}
\textbf{Method} / $\mb{\tau}$\textbf{-Delay} & \textbf{1.90e-02} & \textbf{1.16e-02} & \textbf{7.12e-03} \\
\Xhline{3.5\arrayrulewidth}
MMD-full-traj ($I_s = 50$) & $1.20e\sminus 01 \pm 4.9e\sminus 02$ & $1.02e\sminus 01 \pm 4.1e\sminus 02$ & $8.67e\sminus 02 \pm 3.2e\sminus 02$  \\
\hline
MMD-conditionals & $1.33e\sminus 01 \pm 3.5e\sminus 02$ & $1.42e\sminus 01 \pm 6.1e\sminus 02$ & $1.42e\sminus 01 \pm 4.2e\sminus 02$  \\
\hline
MMD-marginals & $8.20e\sminus 02 \pm 4.5e\sminus 02$ & $1.05e\sminus 01 \pm 3.4e\sminus 02$ & $9.00e\sminus 02 \pm 2.3e\sminus 02$  \\

\Xhline{1.5\arrayrulewidth}
MMD-full-traj ($I_s = 100$) & $1.56e\sminus 01 \pm 8.8e\sminus 02$ & $1.27e\sminus 01 \pm 4.3e\sminus 02$ & $1.06e\sminus 01 \pm 2.8e\sminus 02$ \\ 
\hline
MMD-conditionals & $1.39e\sminus 01 \pm 4.0e\sminus 02$ & $1.68e\sminus 01 \pm 5.7e\sminus 02$ & $1.33e\sminus 01 \pm 4.7e\sminus 02$ \\
\hline
MMD-marginals & $9.75e\sminus 02 \pm 4.7e\sminus 02$ & $9.96e\sminus 02 \pm 3.4e\sminus 02$ & $9.97e\sminus 02 \pm 2.3e\sminus 02$ \\
\Xhline{1.5\arrayrulewidth}
MMD-full-traj ($I_s = 200$) & $2.22e\sminus 01 \pm 1.6e\sminus 01$ & $1.69e\sminus 01 \pm 7.7e\sminus 02$ & $1.51e\sminus 01 \pm 3.8e\sminus 02$ \\ 
\hline
MMD-conditionals & $1.67e\sminus 01 \pm 6.2e\sminus 02$ & $2.28e\sminus 01 \pm 8.3e\sminus 02$ & $2.28e\sminus 01 \pm 3.7e\sminus 02$ \\
\hline
MMD-marginals & $1.16e\sminus 01 \pm 3.1e\sminus 02$ & $1.24e\sminus 01 \pm 2.3e\sminus 02$ & $1.15e\sminus 01 \pm 2.4e\sminus 02$ \\
\Xhline{1.5\arrayrulewidth}
\end{tabular}
}
\caption{\textbf{$L^1$-Accuracy of Learned Force-Law.}
We investigate how the relative $L^1$-errors of the time-dependent marginal
probability densities at times $I_{step}=50,100,200$, $\mbox{epoch} = 5000$,
compare between the learned generative models and the target stochastic
process.  The mean and standard deviation are reported over 5 runs.  SDYN-GANs
training protocols are based on trajectory samples from distributions of the
(i) full trajectory, (ii) conditionals, and (iii) marginals, for details see
Section~\ref{sec_training_protocols}.  The $L^1$-relative-errors of the
empirical radial probability distributions were computed from histograms using
the range $[0,L]$ with $L = 2.5$ and bin-width $\Delta{x} = 0.05$. The total
duration of all trajectories are $t_{n_t}=2.0\times 10^{-2}$ over $n_t=20$
steps with $\Delta{t} = 10^{-3}$.   
}
\label{table_dw2_DNN_L1_err}
\end{table}

We find the generative models learn the force-law with an accuracy of around
$20\%$.  We found the training protocols using the full trajectory and the
marginals performed best, see Table~\ref{table_dw2_DNN_L1_err}.  We remark  
that in theory the large $\tau$ in the infinite time horizon limit would give samples
for $\mb{X}$ from the equilibrium Gibbs-Boltzmann distribution, $\rho(\mb{X}) =
1/Z \exp\left(-U(\mb{X})/k_B{T}\right)$.  The equilibrium distribution only 
depends on the potential $U$ of the force-law.  Interestingly, 
we find in the training there was not a strong dependence in the results 
on the choice of $\tau$.  The results highlight that in
contrast to equilibrium estimators, the SDYN-GANs learning methods allow for
flexibility in estimating the force-law even in the non-equilibrium setting and
from observations over shorter-time trajectories relative to the equilibration 
time-scale.

We found for this task the sampling from the full
trajectory and marginals tended to perform better than the conditionals.  For
the shorter-time durations, this appears to arise from the marginals including
contributions from trajectory fragments that are included in the conditionals
and also trajectory fragments from later parts of the trajectories.  Again,
this provides additional information on the longer-time dynamical responses,
here enhancing learning of the force-laws.  The matching of the marginals 
also may be more stringent during training than the conditionals. Obtaining 
agreement requires more global coordination in the generated trajectories 
so that statistics of later segments of the trajectory match the 
observed target system.  

We remark that for both the conditionals and marginals, the use of shorter duration trajectory 
fragments helps reduce the dimensionality of the samples.  This appears to  
improve in the loss function the ability to make comparisons with 
the observation data.  The results indicate the marginals again may provide
some advantages for training SDYN-GANs when one is able only to probe
trajectory data in fragments over relatively short time durations.

In summary, the results show a few strategies for using SDYN-GANs for learning 
non-linear force-laws and other physical properties from observations of 
the system dynamics.  The samples-based methods of SDYN-GANs allows 
for general generative modeling classes to be used, without a need 
for specification of likelihood functions.  This facilitates the use 
of general generative models for learning
representations of stochastic systems.  The SDYN-GANs approaches also can be
used for non-neural network modeling classes, and combined with other training
protocols, such as data augmentation, further regularizations, and other prior
information motivated by the task and knowledge from the application domain.

\section{Conclusions}
We introduced SDYN-GANs for adversarial learning of generative models for
representing stochastic dynamical systems.  A key feature of our approaches
is that they require only empirical trajectory samples from the candidate generative 
model for comparison with observation data.  This allows for using a wide class of 
generative models without the need for specifying likelihood functions.  
We also developed a class of generative models based on stable $m$-step 
stochastic numerical integrators facilitating learning models for 
long-time predictions and simulations.  We introduced several training 
methods for probing dynamics on different time-scales, utilizing
probabilistic dependencies, and for leveraging different statistical 
properties of the conditional and marginal distributions.  As 
discussed in our results, this provides strategies to reduce 
the dimensionality of the statistics used during training.
This also provides flexibility in how features of the dynamics
are used in learning the generative models.  Our developed SDYN-GANs 
approaches facilitate learning robust stochastic dynamical models 
which can be used in diverse tasks arising in machine learning, 
statistical inference, dynamical systems identification, and 
scientific computation. 
 
\section*{Acknowledgments} 
Authors PJA and CD were supported by grant DOE Grant ASCR PHILMS
DE-SC0019246.  The author PJA was also supported by NSF Grant DMS-1616353.
Author PJA also acknowledges UCSB Center for Scientific Computing NSF MRSEC
(DMR1121053) and UCSB MRL NSF CNS-1725797.  The work of PS is supported by the
DOE-ASCR-funded “Collaboratory on Mathematics and Physics-Informed Learning
Machines for Multiscale and Multiphysics Problems (PhILMs).” Pacific Northwest
National Laboratory is operated by Battelle Memorial Institute for DOE under
Contract DE-AC05-76RL01830.  Author PJA would also like to acknowledge a
hardware grant from Nvidia.

\appendix

\section*{Appendix}

\section{Gradient Equations for $m$-Step Generative Models}
\label{appendix_m_step_recur}

For our $m$-step generative models, the use of direct backpropagation for
book-keeping the forward calculations in generating trajectory samples can result
in prohibitively large computational call graphs.  This is further increased as
the time duration of the trajectories grow.  We derive alternative adjoint
approaches to obtain more efficient methods for loss functions to compute the
gradients needed during training.

For our general $m$-step generative models discussed in Section~\ref{sec_m_step_gradients}, 
we use the formulation
$\mb{f}(\mb{x},\mb{p}) = 0$.  For $m \leq j \leq N$, we have
\begin{eqnarray}
\label{equ_f_app}
[\mb{f}]_{(j,d)} = \left[\mb{X}_j - \bsy{\Psi}_{j-1}(\mb{X}_{j-1},\ldots,\mb{X}_{j-m},\bsy{\omega}_{j-1};\mb{p})\right]^{(d)} = 0.
\end{eqnarray}
The index $(j,d)$ refers to the $j^{th}$ time-step and $d^{th}$ dimension
component associated with $\mb{X}_j^{(d)}$. We use for this the notations
$\mb{X}_j^{(d)} = \left[\mb{X}_j\right]^{(d)} = \mb{X}_{(j,d)}$. The update
map is denoted as $\bsy{\Psi}_{j-1}$. For $j = 0,\ldots,m-1$, we
have 
\begin{eqnarray}
\label{equ_f_init_app}
[\mb{f}]_{(0,d)} = \left[\mb{X}_0 - \mb{x}_0\right]^{(d)} = 0, \;\; 
[\mb{f}]_{(m-1,d)} = \left[\mb{X}_{m-1} - \mb{x}_{m-1}\right]^{(d)} = 0.
\end{eqnarray}
We can then provide the gradients for $\nabla_{X_k}
\bsy{\Psi}_{k},\ldots,\nabla_{X_k} \bsy{\Psi}_{k+m}$ and other terms. 

We derive methods for solving efficiently $J^T\mb{r} = -\mb{f}_p$ 
for $\mb{r}$ where $J = \mb{f}_x$.  For details motivating this formulation, see 
Section~\ref{sec_m_step_gradients}. By using the structure of $J=\mb{f}_x$ from 
equations~\ref{equ_f_app}--~\ref{equ_f_init_app}, 
we can obtain the following set of recurrence equations for the components of 
$\mb{r}$. 
For $0 \leq k \leq N - m$, we have 
\begin{eqnarray}
\label{equ_r1}
r_{(j,d)} f_{(j,d),(k,d')} = r_{(k,d)}\delta_{d,d'} - \left[\nabla_{X_{(k,d')}}\Psi_{(k,d)}\right]r_{(k+1,d)} \cdots
- \left[\nabla_{X_{(k,d')}}\Psi_{(k+m,d)}\right]r_{{(k+m,d)}}
= \frac{\partial \phi}{\partial X_{(k,d')}}.
\end{eqnarray}
For $k = N-\ell,\; 1\leq \ell \leq m-1$, we have 
\begin{eqnarray}
\label{equ_r2}
r_{(j,d)} f_{(j,d),(k,d')} = r_{(k,d)}\delta_{d,d'} - \left[\nabla_{X_{(k,d')}}\Psi_{(k,d)}\right]r_{(k+1,d)} \cdots
- \left[\nabla_{X_{(k,d')}}\Psi_{(k+\ell,d)}\right]r_{{(k+\ell,d)}}
= \frac{\partial \phi}{\partial X_{(k,d')}}.
\end{eqnarray}
For $k = N$, we have 
\begin{eqnarray}
\label{equ_r3}
r_{(j,d)} f_{(j,d),(k,d')} = r_{(N,d)}\delta_{d,d'}
= \frac{\partial \phi}{\partial X_{(N,d')}}.
\end{eqnarray}
This gives an $m^{th}$-order recurrence relation we can use to propagate
values from $N,N-1,\ldots,0$ to obtain the components of $\mb{r}$. 

To obtain $\mb{f}_p$, we have for $m \leq j \leq N$, that 
\begin{eqnarray}
\label{equ_fp1}
[\mb{f}_p]_{(j,d)} = -\nabla_p \bsy{\Psi}_{(j-1,d)}.
\end{eqnarray}
For $0 \leq j \leq m-1$, we have 
\begin{eqnarray}
\label{equ_fp2}
[\mb{f}]_{(j,d)} = \left[\mb{X}_j - \mb{x}_j\right]^{(d)} = 0,\;\; [\mb{f}]_{(j,d),(k,d')} =\delta_{(j,d),(k,d')},\;\; 
[\mb{f}_p]_{(j,d)} = -\frac{\partial \left[\mb{x}_j\right]^{(d)}}{\partial p}.
\end{eqnarray}
We compute the gradients needed for training by using 
\begin{eqnarray}
\label{equ_gp1}
\nabla_p \tilde{g} = \tilde{\mathbf{g}}_p - \mathbf{r}^T\mathbf{f}_p.
\end{eqnarray}
To compute $\mb{r}$, we solve for each sample $\mb{X}^{[i]}$ of the stochastic
trajectories the $m^{th}$-order recurrence relations given in
equations~\ref{equ_r1}--~\ref{equ_r3}.  The $\tilde{\mathbf{g}}_p = \partial
\tilde{\mb{g}}/\partial p$ and $\mathbf{f}_p$ are computed using
equations~\ref{equ_fp1}--~\ref{equ_fp2}. The final gradient is computed by
averaging using $\nabla_p \bar{g} = \frac{1}{M} \sum_{1=1}^M \nabla_p
\tilde{g}^{[i]}$ over the $M$ samples of the stochastic trajectories indexed by
$[i]$.  There are also ways to obtain additional efficiencies by using
structure of the specific loss functions, as we discuss in
Section~\ref{sec_backprop_loss}.  We use
equations~\ref{equ_f_app}--~\ref{equ_gp1} to develop efficient methods
for computing the gradients of our $m$-step generative models.

\newpage
\clearpage
\bibliography{paper_database}

\end{document}